\documentclass[10pt, conference, letterpaper]{IEEEtran}
\IEEEoverridecommandlockouts
\usepackage{cite}
\usepackage{amsmath,amssymb,amsfonts}
\usepackage[ruled,linesnumbered]{algorithm2e}
\usepackage{algpseudocode}
\usepackage{textcomp}
\usepackage{xcolor}

\makeatletter
\renewcommand\paragraph{\@startsection{paragraph}{4}{\z@}%
	{1.5ex plus .2ex minus .3ex}%
	{-0em}%
	{\normalsize\bf}}
\makeatother

\ifCLASSINFOpdf
    \usepackage[pdftex]{graphicx}
    \DeclareGraphicsExtensions{.pdf,.jpeg,.png}
\else
   \usepackage[dvips]{graphicx}
  \DeclareGraphicsExtensions{.eps}
\fi
\graphicspath{{figs/}}
\usepackage[caption=false]{subfig}

\def\BibTeX{{\rm B\kern-.05em{\sc i\kern-.025em b}\kern-.08em
    T\kern-.1667em\lower.7ex\hbox{E}\kern-.125emX}}
\begin{document}

\title{Informative Path Planning for Mobile Sensing with Reinforcement Learning}

 \author{\IEEEauthorblockN{Yongyong Wei, Rong Zheng}
 \IEEEauthorblockA{Department of Computing and Software,
 McMaster University\\
 1280 Main St. W., Hamilton, ON, Canada \\
 Emails: \{weiy49, rzheng\}@mcmaster.ca}
 }

\maketitle

\begin{abstract}
Large-scale spatial data such as air quality, thermal conditions and location signatures play a vital role in a variety of applications. Collecting such data manually can be tedious and labour intensive. With the advancement of robotic technologies, it is feasible to automate such tasks using mobile robots with sensing and navigation capabilities. However, due to limited battery lifetime and scarcity of charging stations, it is important to plan paths for the robots that maximize the utility of data collection, also known as the {\it informative path planning} (IPP) problem. In this paper, we propose a novel IPP algorithm using reinforcement learning (RL). A constrained exploration and exploitation strategy is designed to address the unique challenges of IPP, and is shown to have fast convergence and better optimality than a classical reinforcement learning approach. Extensive experiments using real-world measurement data demonstrate that the proposed algorithm outperforms state-of-the-art algorithms in most test cases. Interestingly, unlike existing solutions that have to be re-executed when any input parameter changes, our RL-based solution allows a degree of transferability across different problem instances.
\end{abstract}


\section{Introduction}
\label{sect:intro}
A wide range of applications rely on the availability of large-scale spatial data, such as water and air quality monitoring, precision agriculture, WiFi fingerprint based indoor localization, etc. One common characteristic of these applications is that the data to be collected are location dependent, and time consuming to obtain if done manually. Over the last two decades,  wireless sensor networks (WSN)~\cite{akyildiz2002wireless} have been extensively investigated as a means of continuous environment monitoring. To exploit mobility, WSN with mobile elements~\cite{di2011data} has also been considered. While individual sensor devices are typically at low costs, deploying and maintaining a large-scale WSN incur high capital and operational expenses.

For one-time or infrequent spatial data collection, robotic technologies offer a viable alternative to fixed deployments~\cite{wang2007robot}. A robot equipped with sensing devices can be controlled to traverse a target area and collect environmental data along its path. 
Although utilizing robots for spatial information gathering can significantly reduce human efforts, they are battery powered and have limited life time. Given a budget constraint (e.g., maximum travel distance or time), it is important to plan motion trajectories for the robots such that the state of the environment can be accurately estimated with the sensor measurements. 

In this work, we model the distribution of spatial data in a target area as a \textit{Gaussian Process} (GP)~\cite{williams2006gaussian}. 
GPs are versatile in that by choosing  appropriate kernel functions, it can be used to model processes of different degrees of smoothness. In prediction, besides the predicted values, uncertainties (variances) are also provided. Based on GPs, in~\cite{guestrin2005near}, \textit{mutual information} (MI) is proposed as a criteria to measure the informativeness for sensor placement. In~\cite{singh2007efficient,binney2010informative,meliou2007nonmyopic}, MI is used to measure the informativeness of a path when data are collected by a robot following the path. The problem of finding the most informative path from a pre-defined start location to a terminal location subject to a budget constraint is called \textit{informative path planning} (IPP).  

In general, IPP problems are formulated on graphs~\cite{meliou2007nonmyopic,binney2010informative,binney2012branch}, with vertices representing way-points and edges representing path segments. The utility\footnote{In this work, we use the terms utilty, reward and informativeness exchangeably.} of a path can be associated with the vertices, edges or both on the path. In the special case where utility is limited to vertices and is additive, the IPP problem degenerates to the well-known Orienteering Problem (OP), which is known to be NP hard~\cite{vansteenwegen2011orienteering}. Existing solutions to IPP mostly adopt heuristics based search strategies such as greedy search~\cite{yu2014informative} and evolutionary algorithms~\cite{wei2018informative,viseras2016planning}. These heuristics often suffer from inferior performance. Furthermore, even with small changes in input parameters, the heuristic solution needs to be re-executed. 

In this paper, a novel \textit{reinforcement learning} (RL) algorithm is proposed to solve the IPP problem. Specifically, we model IPP  as a sequential decision process. Given the start vertex on the IPP graph, a path is constructed sequentially by appending the next way-point vertex. With reinforcement learning, the total rewards of the generated paths are expected to improve gradually. 

Compared with conventional RL tasks, IPP poses non-trivial challenges. The available actions depend on the current position of the agent on the graph, since it can only choose among adjacent vertices as the next step. Furthermore, the reward of an action depends on past actions. For instance, re-visiting a vertex can lead to less but non-zero reward. Lastly, eligible paths (states) are constrained by the budget and the pre-defined terminal vertex. As a result, RL needs to be tailored to the problem setting. We adopt a recurrent neural network (RNN) based Q-learning structure, and select feasible actions using a mask mechanism. In order to improve learning efficiency, a constrained exploration and exploitation strategy is devised. Such a strategy allows looking ahead and restricting to valid paths that can terminate at the specified vertex within budget constraint.  

To evaluate the proposed approach, we consider the task of WiFi Radio Signal Strength (RSS) collection in indoor environments. WiFi RSS measurements are commonly used in fingerprint-based indoor localization solutions~\cite{wu2012will,yang2012locating,li2017turf}. Real data have been collected from two areas. In total, 20 different configurations (different start/terminal vertices, or budget constraints) have been evaluated. Among them, the RL based IPP algorithm outperforms state-of-the-art methods in 17 configurations with higher informativeness. Furthermore, we find that when the change in configuration is small, transfer learning from a pre-trained model can greatly improve the convergence speed on a new problem instance. 

The rest of this paper is organized as follows.  In Section~\ref{sect:rw}, related work to IPP and a background of RL are introduced briefly. The IPP problem is formally formulated in Section~\ref{sect:probform}.  We present the proposed solution in Section~\ref{sect:ps}. Experimental results are shown in Section~\ref{sect:pe}, and we conclude our work in Section~\ref{sect:conclustion}.


\section{Related Work and Background}
\label{sect:rw}
In this section, representative solutions to IPP are reviewed first. A brief background of RL is then presented, with a focus on the Q-learning approach. We also review two recent works that attempt to solve the combinatorial optimization problem with RL.

\subsection{Existing solutions to IPP}
In~\cite{wei2018informative}, IPP has been shown to be NP-hard. A greedy algorithm and a \textit{genetic algorithm} (GA) are investigated. Experiments show that GA achieves a good trade-off between computation complexity and optimality. In~\cite{viseras2016planning}, another evolution based path planning approach is proposed with ant colony optimization. In~\cite{macdonald2019active}, the path planning process is modeled as a control policy and a heuristic algorithm is proposed by incrementally constructing the policy tree. 

Several  algorithms decompose the optimization problem into  subset selection and path construction. The main intuition is that once the subset of vertices are determined, a TSP solver can be used to construct a path with the minimum cost. In~\cite{arora2017randomized}, vertices are randomly added or removed, and a TSP solver is used to maintain the path. Similarly, in~\cite{ma2017informative}, way-points are added incrementally and a TSP solver is used to determine the traversing order. Such approaches usually assume that each selected vertex can only be visited once (due to TSP) and the reward is accumulated only from  the selected vertices. In IPP applications, such assumptions do not generally hold since robots can continue sensing the environment while travelling along the path. Furthermore, a vertex can be visited multiple times and rewards can still be obtained, particularly when MI is used as the criteria of informativeness. 

Another line of IPP algorithms are based on the recursive greedy  (RG) algorithm proposed for OP~\cite{chekuri2005recursive}. RG is an approximate algorithm.  The basic idea is to consider all possible combinations of intermediate vertices and budget splits, and then it is recursively applied on the smaller sub-problems. IPP with RG can be found in ~\cite{singh2007efficient,binney2010informative}. In order to reduce computation complexity, in~\cite{singh2007efficient}, the authors propose spatial decomposition to create a coarse graph by grouping the vertices. Unfortunately, doing so can compromise the approximation guarantee of the original algorithm.

Most of the above mentioned algorithms suffer from a limited performance in terms of optimality. On the other hand, although RG has an approximation guarantee, it is not practical on large graphs or when the budgets are large due to its complexity.
\subsection{Reinforcement Learning}
Under the framework of RL~\cite{kaelbling1996reinforcement,sutton2018reinforcement},  an agent interacts with the environment through a sequential decision process, which can be described by a Markov Decision Process (MDP) $<\mathcal{S},\mathcal{A},\mathcal{T},\mathcal{R}>$, where
\begin{itemize}
    \item $\mathcal{S}$ is a finite set of states;
    \item $\mathcal{A}$ is a finite set of actions;
    \item $\mathcal{T}$ is a state transition function\footnote{In this work we consider deterministic transitions.} defined as $\mathcal{T}: \mathcal{S} \times \mathcal{A} \xrightarrow{} \mathcal{S} $;
    \item $\mathcal{R}$ is a reward function defined as $\mathcal{R} : \mathcal{S} \times \mathcal{A} \xrightarrow{} \mathbb{R}$, where $\mathbb{R}$ is a real value reward signal. 
\end{itemize}

To solve the MDP with RL, a policy $\pi$ is required to guide the agent towards decision making. The policy can be deterministic or stochastic. A deterministic policy is defined as $\pi(s): \mathcal{S} \xrightarrow{} \mathcal{A}$, i.e., given the state, the policy outputs the action to take for the following step. 

At each time step $t$, the environment is at a state $s_t \in \mathcal{S}$. The agent makes a decision by taking an action $a_t = \pi(s_t) \in \mathcal{A}$. It then receives an immediate reward signal $r_t$ and the state moves to
$s_{t+1} \in \mathcal{S}$. The goal of RL is to find a policy $\pi$ such that the total future reward 
\begin{equation}
    R_t = r_t + \gamma r_{t+1} +...+ \gamma ^{T-t}r_T 
\end{equation}
is maximized, where $\gamma \in [0,1]$ is a discount factor controls the priority of step reward and $T$ is the last action time. 

There are two main approaches to find the desired policy $\pi$, namely the policy-based and the value-based approaches. The policy-based approach (e.g., policy gradient) aims to directly optimize the policy and output the action (or action distribution for non-deterministic policy) given an input state, while the value-based approach (e.g., Q-learning) is indirect. The insight is to predict the total future reward given an input state or a state-action pair, the agent can then make decisions through the predicted reward.  

We consider the Q-learning approach in this work. Specifically, Q-learning aims to learn a function $\mathcal{Q}: \mathcal{S} \times \mathcal{A} \xrightarrow{} \mathbb{R}$, with $\mathcal{Q}(s,a)$ representing the  total future reward by taking the action $a$ from state $s$. The  policy given $\mathcal{Q}$ can then be formulated as 
\begin{equation}
    \pi(s) = \arg\max_a \mathcal{Q}(s,a).
\end{equation}

In practice, the Q-function is usually approximated with a neural network $\mathcal{Q}_\theta(s,a)$,  which is known as DQN~\cite{mnih2013playing}. The network is optimized in an iterative way by minimizing the temporal difference with a loss function defined as 
\begin{equation}
\label{eq:tempoloss}
    L(\theta) = \large( \mathcal{Q}(s_t, a_t) - (r_t + \gamma \underset{a}{max} \mathcal{Q}(s_{t+1},a)) \large)^2.
\end{equation}

There are many variants and techniques for Q-learning models and training methodology~\cite{van2016deep,wang2015dueling,schaul2015prioritized}. We only cover the basic background here due to space limitation and Q-learning itself is not a part of our contribution. Most of these techniques can be applied directly in our proposed method.

In recent years, RL with neural network has been applied to solve combinatorial optimization problems. In~\cite{bello2016neural}, the authors consider TSP and utilize a pointer network to predict the distribution of vertex permutations. Negative tour lengths are used as reward signals, and parameters of the neural network are optimized using the policy gradient method. Experiments show that neural combinatorial optimization achieves close to optimal results on 2D Euclidean graphs.

In~\cite{khalil2017learning}, a Q-learning approach is presented to solve the combinatorial optimization problems on graphs. A graph embedding technique is desinged for graph representation, and solutions are greedily constructed with Q-learning. The effectiveness of the approach is evaluated on Minimum Vertex Cover, Maximum Cut and TSP.

Both~\cite{khalil2017learning} and~\cite{bello2016neural} assume complete graphs. In contrast, presence of obstacles in spatial areas implies that the resulting graphs have limited connectivity. Furthermore, as discussed previously, IPP is fundamentally a harder problem than TSP, and in some cases TSP is a sub-process for some IPP solutions. In this paper, we show how RL can be applied in the IPP context.

\section{Problem Formulation}
\label{sect:probform}

Since IPP is defined on graphs, the target area needs first to be converted to a graph. Points of Interests (PoIs) in the area can be seem as vertices, and an edge exists if two vertices are reachable. 

\subsection{General Path Planning with Limited Budget}
We define the graph-based general path planning problem using a five-tuple $<G,v_s,v_t,f(\mathcal{P}),B>$. Specifically,
\begin{itemize}
    \item $G = (V,E)$ is the graph. Each $v \in V$ is associated with a physical location $\mathbf{x}$. For each $e \in E$, there is a corresponding cost $c_e$ (e.g., the length of the edge) for travelling along the edge.
    \item $v_s, v_t \in V$ is the specified start and terminal vertex, respectively. 
    \item  A valid path\footnote{In graph theory, a path is defined as a sequence of vertices and edges without repeated vertices or edges. To be consistent with existing IPP literature, we  allow repetition of vertices on a path, the equivalent of a walk in graph theory.} is denoted by  $\mathcal{P} = [v_s,v_1,...,v_k,v_t]$, and its reward is denoted by  $f (\mathcal{P})$. 
    \item $B$ is the total budget available for the path. 
\end{itemize}

The cost of  $\mathcal{P}$ is the sum of edge costs along the path,
\begin{equation}
    C(\mathcal{P}) = \sum_{i=1}^{n-1} c_{(v_i(\mathcal{P}) ,v_{i+1}(\mathcal{P}) )},
\end{equation}
where $v_i(\mathcal{P}) $ is the $i$-th vertex in $\mathcal{P}$ and $(v_i(\mathcal{P}),v_{i+1}(\mathcal{P}) )$ represents the corresponding edge. The objective is to find the optimal path that satisfies
\begin{equation}
\label{eq:maximization}
\mathcal{P}^* = \arg \max_{\mathcal{P} \in \Psi} f(\mathcal{P})~s.t.~ C(\mathcal{P}) \leq B,
\end{equation}
where $\Psi$ is the set of all paths in $G$ from $v_s$ to $v_t$.

One classic variant of the general path planning formulation is OP~\cite{golden1987orienteering,vansteenwegen2011orienteering,gunawan2016orienteering}.  In OP, each vertex is associated with a reward and the goal is to find a subset of vertices to visit so as to maximize the collected reward within a budget constraint. When $f(\mathcal{P})$ is submodular or correlated, it is also known as the submodular orienteering problem (SOP)~\cite{chekuri2005recursive} or correlated orienteering problem (COP)~\cite{yu2014correlated}.

\subsection{Informative Path Planning}
IPP is a specific case of the general path planning problems where the reward of a path is defined by the informativeness of data collected along the path.  In information theory, informativeness can be measured through MI~\cite{singh2007efficient,binney2010informative,cao2013multi,ma2017informative}. Next, we present the calculation of $f(\mathcal{P})$ for IPP based on GPs and MI. Detailed mathematical background of GPs can be found in~\cite{williams2006gaussian}.

Assume the data to be collected can be modeled by a GP. Thus, for each $v \in V$ at a physical location $\mathbf{x}$, the corresponding data $y_v$ (e.g., temperature, humidity, etc.) is a Gaussian distributed random variable, and the variables $\mathbf{y}_V$ at all the locations of $V$ follow a joint multivariate Gaussian distribution,
\begin{equation*}
\mathcal{N} 
\begin{pmatrix}
\begin{bmatrix}
m(\mathbf{x}_1)\\
\vdots\\
m(\mathbf{x}_n)
\end{bmatrix}\!\!\ ,&
\begin{bmatrix}
k(\mathbf{x}_1,\mathbf{x}_1) & ... & k(\mathbf{x}_1,\mathbf{x}_n)\\
\vdots & \ddots & \vdots\\
k(\mathbf{x}_n,\mathbf{x}_1) & ... & k(\mathbf{x}_n,\mathbf{x}_n)
\end{bmatrix}
\end{pmatrix} ,
\end{equation*}
where $m(\mathbf{x})$ is the mean function and $k(\mathbf{x}_p,\mathbf{x}_q)$ is the kernel, and $n = |V|$ is the total number of vertices. For simplicity, we denote the multivariate Gaussian distribution by $\mathcal{N}(m(X_V),\Sigma_V)$, where $X_V$ is a $n \times 2$ matrix for the locations of $V$ and $\Sigma_V$ is the $n \times n$ covariance matrix as defined by the above kernel function $k$.

The differential entropy (also referred to as continuous entropy)~\cite{ahmed1989entropy} of $\mathbf{y}_V$ is
\begin{equation}
    H(\mathbf{y}_V) = \frac{1}{2} ln |\Sigma_V| + \frac{n}{2}(1 + ln (2\pi)).
\end{equation}

Given $\mathcal{P} = [v_s,v_1,...,v_k,v_t]$, suppose data are going to be collected by an agent along the path every $d$ meter interval (depends on the traveling speed and sampling frequency). The sample locations can be easily calculated with the positions of the vertices. We denote all the sample locations as $X_S$ and the corresponding measurements as $\mathbf{y}_S$. The posterior distribution of $\mathbf{y}_V$ given $\mathbf{y}_S$ is $\mathcal{N}(\mathbf{\mu}',\Sigma')$, where
\begin{multline}
\label{eq:gp_mu}
\mathbf{\mu}' = m(X_V) + K(X_V,X_S)(K(X_S,X_S)+\sigma_n^2 I)^{-1}\\(\mathbf{y}_S - m(X_S)),
\end{multline}
\begin{multline}
\label{eq:gp_sigma}
\Sigma' = K(X_V,X_V) - K(X_V,X_S)(K(X_S,X_S) + \sigma_n^2 I )^{-1}\\K(X_S,X_V).
\end{multline}
Here $\sigma_n$ represents the noise variance of the underlying GP, and $K(X_V,X_S)$ is the kernel matrix generated by $k(\cdot,\cdot)$ with pair-wise entries in $X_V$ and $X_S$. The conditional differential entropy is then given by 
\begin{equation}
    H(\mathbf{y}_V |\mathbf{y}_S) = \frac{1}{2} ln |\Sigma '| + \frac{n}{2}(1 + ln (2\pi)).
\end{equation}
The MI based reward can be calculated with 
\begin{equation}
f(\mathcal{P}) = MI({\mathbf y}_V;{\mathbf y}_S) = H({\mathbf y}_V) - H({\mathbf y}_V|{\mathbf y}_S).
\end{equation}

Note that since the differential entropy only depends on the kernel matrix (i.e, the kernel function and the locations), reward can be calculated analytically without travelling along the actual path and taking real measurements. That is why it is possible for offline path planning.

However, the kernel function $k(\cdot,\cdot)$ usually has some hyperparameters which may not be known in advance. Thus, pilot data are needed to learn these hyperparameters~\cite{
binney2010informative,guestrin2005near,chekuri2005recursive}. Given a small set of pilot data $D = (X_D, \mathbf{y}_D)$ collected in advance at locations $X_D$ with measurements $\mathbf{y}_D$, the reward can be calculated with 
\begin{equation}
\label{eq:fD}
f_D(\mathcal{P}) = MI({\mathbf y}_V;{\mathbf y}_S \cup \mathbf{y}_D) = H({\mathbf y}_V) - H({\mathbf y}_V|{\mathbf y}_S \cup \mathbf{y}_D).
\end{equation}

Given the input as $<G,v_s,v_t,f(\mathcal{P}),B>$, one naive approach is to enumerate all the valid paths from $v_s$ to $v_t$ and choose the path with the highest $f(\mathcal{P})$. However, since the problem is NP-hard, brute force search is not computationally feasible in practice.

\section{Proposed Solution}
\label{sect:ps}

In this section, we present the proposed solution with a Q-learning approach. Related concepts are defined first. Then we present the overview and details of each component.

\begin{figure}[!t]
\centering
\includegraphics[width=0.45\textwidth]{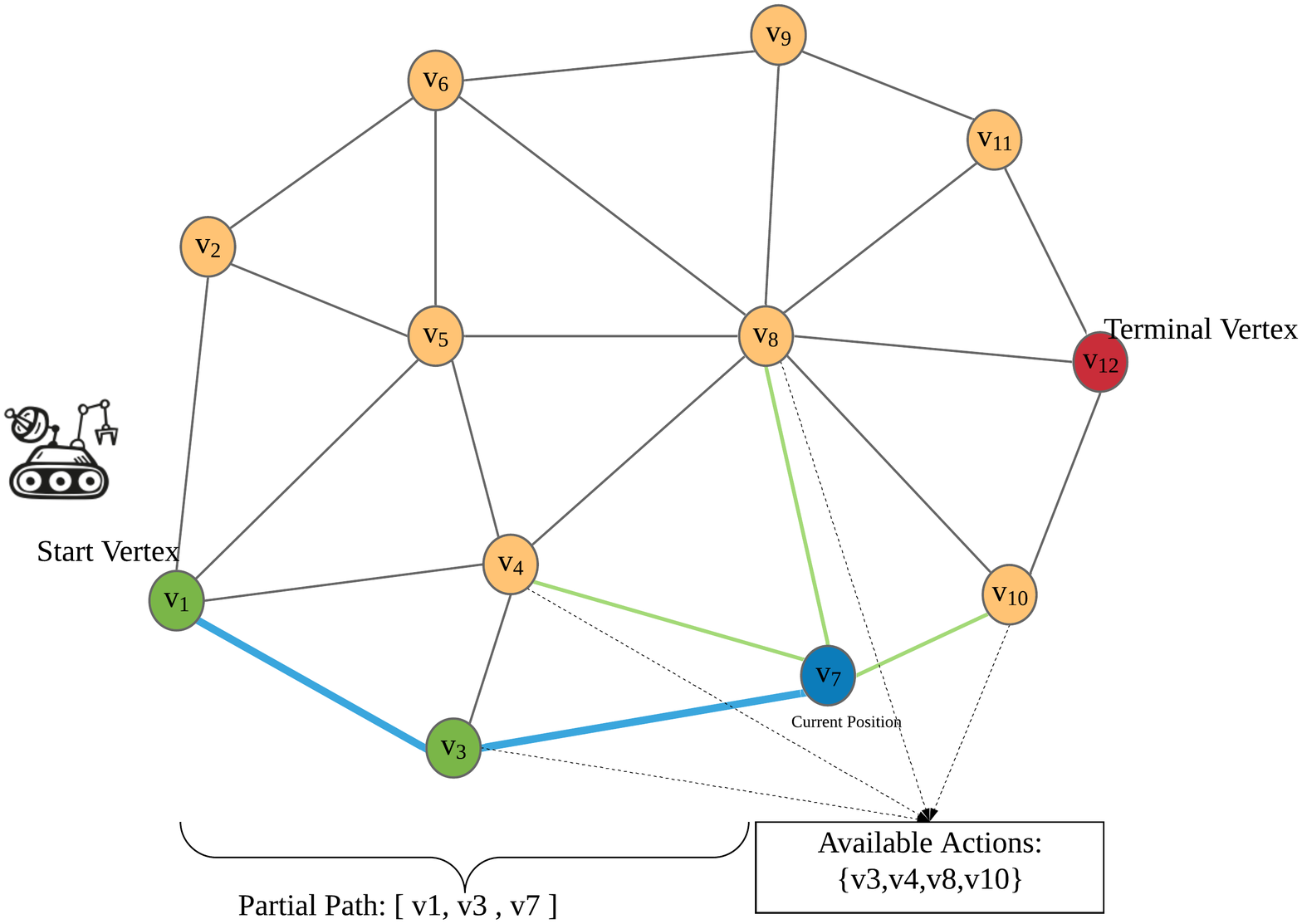}
\caption{Sequential Decision Process for IPP}
\label{fig:sdp}
\end{figure}
It is straightforward to view IPP as a sequential decision problem. Specifically, suppose an agent is exploring solutions in $G$ from $v_s$ to $v_t$, with a budget $B$. As shown in Fig.~\ref{fig:sdp}, we denote the vertices traversed by the agent as the partial path $\mathcal{P}_{p}$. Initially, $\mathcal{P}_{p} = [v_s]$. In subsequent steps,  available actions for the agent are the adjacent vertices of the last vertex in $\mathcal{P}_{p}$, i.e., the current position of the agent. Once the agent decides which action to take according to some policy $\pi$, the action (vertex) is appended to the partial path, and a corresponding immediate reward is sent to the agent. The process repeats until the budget is exhausted or the agent successfully reaches $v_t$.  
We summarize the corresponding RL concepts in the context of IPP as follows,
\begin{itemize}
    \item \textbf{Agent and Environment}: An agent is a robot at $v_s$ and moves along the edges. The environment is a simulator based on the input graph. 
    \item \textbf{State}: Many RL solutions such as~\cite{mnih2013playing} encode the states with pixel level images and use convolutional neural network (CNN) for an end to end learning.  For IPP, since it is defined on a graph, it is not necessary to  use CNN.  Instead, we define the state with $\mathcal{P}_p$ and state transition means appending a vertex  to $\mathcal{P}_p$.
    \item \textbf{Action}: Action means selecting which available vertex to go for the next step. The available actions (the next way-point to visit) vary significantly when the agent moves to a new vertex, depending on the connectivity of the graph $G$.
    \item \textbf{Reward}: Reward is a numerical value given to the agent by the environment after it takes an action. The rewards are expected to link to the optimization goal, i.e, maximize the informativeness of the path.  
    \item \textbf{Episode}: Each episode represents the process to construct a trial path starting from $v_s$ until the budget is used up or reaches $v_t$.  The agent is expected to reach the terminal vertex within the budget.
\end{itemize}

\subsection{Solution Overview}
\begin{figure}[!t]
\centering
\includegraphics[width=0.45\textwidth]{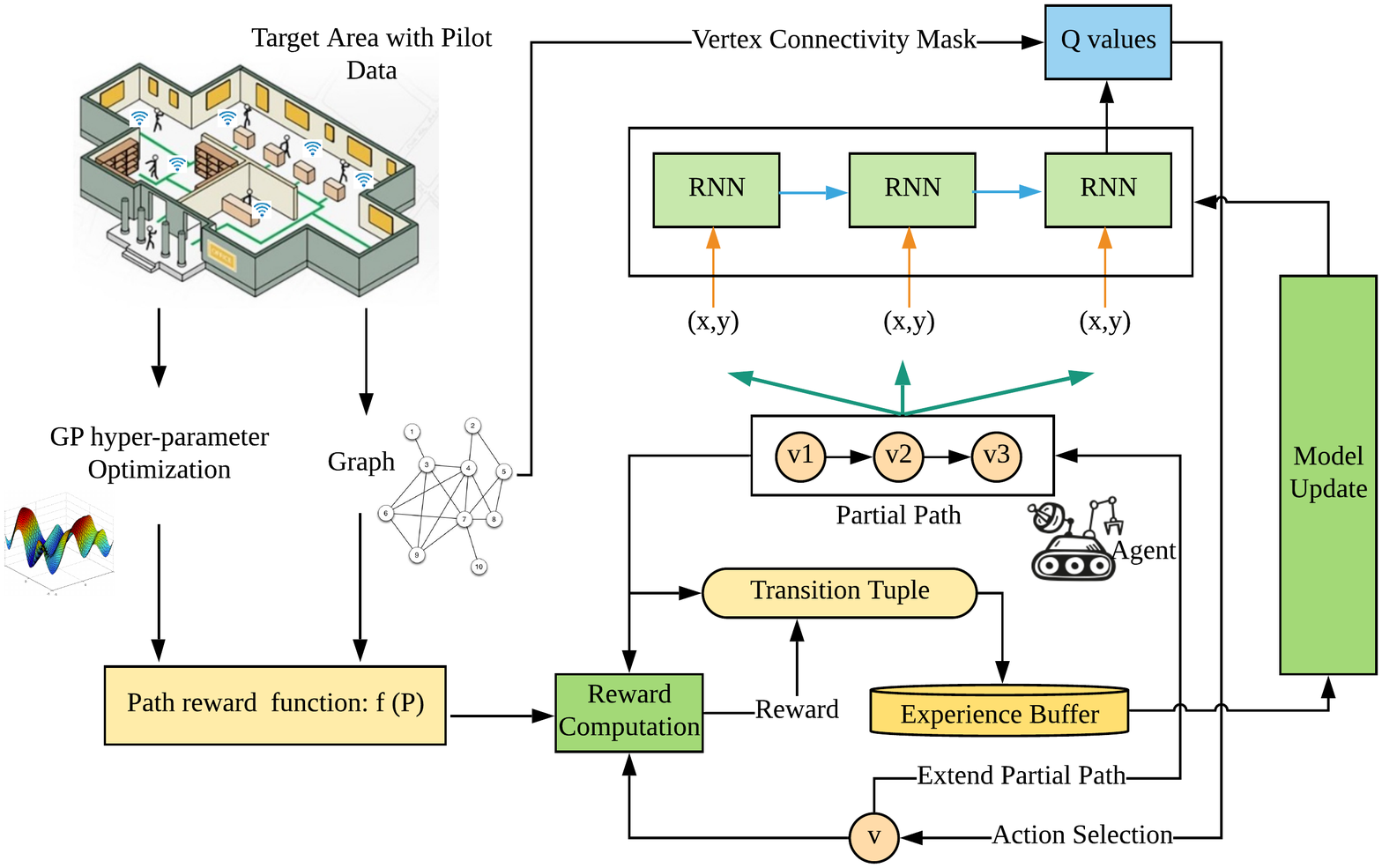}
\caption{Solution overview with Reinforcement Learning}
\label{fig:overview}
\end{figure}
Fig.~\ref{fig:overview} shows the overall architecture of the solution. The  input is a target area with a small amount of pilot data. The area is discretized into a graph. As mentioned previously, the data to be collected are spatially correlated. A GP regression model is fitted and optimized with the pilot data to estimate the hyperparameters. Once the hyperparameters are estimated, the reward function $f_D(\mathcal{P})$ defined in~(\ref{eq:fD}) is determined, which can be used to calculate the step reward for the agent.

We utilize a \textit{Recurrent Neural Network} (RNN)  as the Q-value approximator, since future rewards (Q-values) depend on all the visited vertices. Meanwhile, for each input state, we bind a Q-value for every vertex in the graph, even if it is not a direct neighbor to the last vertex of $\mathcal{P}_{p}$. The Q-values are then masked with the connectivity of the graph to filter out those non-reachable vertices. For each epoch, the agent starts from $v_s$ and select actions with the $\epsilon$-greedy policy based on the output Q-values. Reward is calculated with $f(\mathcal{P})$ and state transition tuples are added to the experience buffer. For each step, a batch of transition tuples are sampled from the buffer, which are utilized to update the model's parameter by minimizing the temporal difference in (\ref{eq:tempoloss}).

Next, we give a detailed description to some of the key components.

\subsection{Constrained Exploration and Exploitation}
\label{sect:cee}
One major obstacle in applying RL to IPP is the \textit{Constrained Terminal State}. Any valid path should start from $v_s$ and terminate at $v_t$, and also satisfy the budget constraint. We design a novel constrained exploration and exploitation strategy that can reduce the computation complexity in finding valid and high-reward paths. 

\subsubsection{Exploration}
Let $\mathcal{P}_{p} = [v_s,...,v_k]$ and $v_k$ is the agent's current location. The set of available actions $\mathcal{A}$ to the agent is given by the neighborhood of $v_k$, i.e.,
\begin{equation}
    \mathcal{A}(\mathcal{P}_{p}) = N(v_k) = \{x \in V(G): (v_k,x) \in E(G)\}.
\end{equation}
Among $\mathcal{A}$ , we denote valid actions that have chances to reach $v_t$ within the budget as $\mathcal{A}'$, 
\begin{multline}
\label{eq:aprime}
    \mathcal{A}'(\mathcal{P}_{p}) = \{x \in \mathcal{A}(\mathcal{P}_{p}): \textrm{Length}(v_k,x) 
    \\+ \textrm{ShortestPath}(x,v_t) \le B_r \},
\end{multline}
where $B_r$ is the remaining budget that can be calculated by subtracting the length of $\mathcal{P}_{p}$ from $B$, and ShortestPath$(x, v_t)$ denotes the least cost path from vertex $x$ to $v_t$. 

If the agent randomly chooses an action from $\mathcal{A}'$ at each step, it can be guaranteed to reach $v_t$ within the budget. Furthermore, among $\mathcal{A}'$, it is possible that some vertices have been visited previously. For IPP, our exploration strategy is to randomly select a vertex that has not been visited if exists, otherwise randomly select a vertex from $\mathcal{A}'$.  Note that $\mathcal{A}'$ changes with the states. In other words, the valid actions are updated step-wise.

\subsubsection{Exploitation}
Through controlling the exploration actions, the agent is guaranteed to reach $v_t$. However, when actions are generated through exploitation with the maximum predicted Q-value, they may be invalid. This is particularly the case in the initial stage when the predicted  Q-values are not accurate. Again, the shortest path  is utilized to identify such actions. If the remaining budget is not sufficient to cover the selected action and the shortest path thereafter, the episode is terminated immediately and a penalty reward is triggered. 

\subsection{Reward Mechanism}
For each action, the environment provides an immediate reward signal and transits to the next state. A simulator is created based on the input graph. 

The reward of taking action $a \in \mathcal{A}'(\mathcal{P}_{p}) $ is defined as
\begin{equation}
\label{eq:actionreward}
    r(\mathcal{P}_p,a) = f(\mathcal{P}_{p} + [a]) - f(\mathcal{P}_{p}),
\end{equation}
In such a way,  the reward in each step adds up to the reward of the constructed path at the last step.

\begin{algorithm}
%
\SetAlgoLined
\caption{State Transition and Reward Mechanism}
\label{alg:strm}
\SetKwInOut{Input}{Input}
\SetKwInOut{Output}{Output}
\SetKwRepeat{Do}{do}{while}
\Input{$<G,v_s,v_t,f(\mathcal{P}),B>$,  $\mathcal{P}_{p}$, $R_{c}, a$}
\Output{Transition Tuple $<s, a, r, s',\textrm{IsDone} >$}

    $s = \mathcal{P}_p$ \\
    $v_k$ = last vertex of $\mathcal{P}_p$ \\
    \uIf{$B_r \geq \textrm{Length}(v_k,a) + \textrm{ShortestPath}(a,v_t)$}
    {
        calculate $r$ according to (\ref{eq:actionreward}) \\
        $\textrm{IsDone} = \textrm{False}$ \\
        \uIf{$a == v_t$}{$\textrm{IsDone} = \textrm{True}$}
         $s' = \mathcal{P}_{p}+[a]$\\
        $\mathcal{P}_{p} = s'$\\
        $R_c = R_c + r$
    }
    \uElse{
         $\textrm{IsDone} = \textrm{True}$ \\
         $r=-1.0 \times R_c$ \\
         $s' = \mathcal{P}_{p}$
    }
Return $<s,a,r,s',\textrm{IsDone}>$
\end{algorithm}

When action $a$ violates the budget constraint, we signal a penalty reward to the agent to discourage such an action. Specifically, a variable $R_c$ is used to track the cumulative reward obtained. Once the budget constraint is violated, the reward perceived becomes the negative of $R_c$. Therefore, any invalid path eventually leads to a zero reward (except the initial reward from the pilot data). The state transition and reward mechanism in one step are outlined in Algorithm~\ref{alg:strm}. The procedure returns a transition tuple $<s, a, r, s',\textrm{IsDone} >$, namely, upon taking action $a$ from state $s$ the agent gets a reward $r$, the state transits to $s'$, and IsDone means whether the action terminates the episode. The transition tuple is stored in an experience buffer, which is the input for Q-network training.

\subsection{Q-learning Network}
The Q-learning network is used to predict the Q-values, for the agent to make better decisions. We adopt a RNN based neural network since the input is a sequence. Given $\mathcal{P}_p$, the input to the RNN is the corresponding 2D location coordinates for each vertex in $\mathcal{P}_p$. The output of the last cell is a Q-value vector $Q_o$ with length $|V|$.

On the other hand, since the graph may not be fully connected and the predicted Q-values are only valid for the adjacent vertices, we define a mask vector $Q_{m}$ of length $|V|$ as 
\begin{equation}
    Q_{m}[i] = \left\{
        \begin{array}{ll}
            0 & \quad v_i \in N(v_k) \\
            M & \quad else
        \end{array}
    \right.
\end{equation}
where $M$ is a predefined large negative number as a penalty reward and $v_k$ is the current position. Therefore, the final masked output of the Q-values are $Q_o + Q_{m}$.

\subsection{Learning and Searching Algorithm}
Based on the Q-network, the agent uses an $\epsilon$-greedy  policy to explore the solution space, with the proposed constrained exploration and exploitation strategy in Section~\ref{sect:cee}. The state transition tuples from Algorithm~\ref{alg:strm} are cached in an experience buffer $\mathcal{M}$, and network parameters are trained based this memory buffer. However, when neural network is used as the function approximator, there is no theoretical convergence guarantee for Q-learning~\cite{mnih2013playing}. Further, with gradient descent based optimization, the final model may stuck at local optima. Thus, it is possible that paths sampled during the training stage may have a larger reward than the paths generated by the final Q-network. We utilize a learning and searching strategy similar to the ``Active Search'' in~\cite{bello2016neural}. For every $K$ iterations, a path is constructed with greedy search according to the Q-network, and we keep track of the best path ever seen as the final solution.

\begin{algorithm}
%
\SetAlgoLined
\caption{Learning and Searching Algorithm}
\label{alg:ls}
\SetKwInOut{Input}{Input}
\SetKwInOut{Output}{Output}
\SetKwRepeat{Do}{do}{while}
\Input{$<G,v_s,v_t,f(\mathcal{P}),B>$, RNN Q-network $\mathcal{Q}$}
\Output{Best Path Found}

Initialize the experience  buffer $\mathcal{M}$\\
Initialize the best path and reward as $\mathcal{P}_b = None,R_b = 0$ \\
\For{episode  e $\gets$ 1 to N}{
    Initialize $\mathcal{P}_p = [v_s]$ \\
    \For{step $t \gets 1$ to T}{
        With probability $\epsilon$ select an action $a \in \mathcal{A}'$\\
        Otherwise select $a = \arg \max_{a}\mathcal{Q}(\mathcal{P}_p,a)$ \\
        Get transition tuple from Algorithm \ref{alg:strm} and store to $\mathcal{M}$ \\

        \uIf{terminates}{
            \uIf{$f(\mathcal{P}_p) > R_b$}{
                $\mathcal{P}_b = \mathcal{P}_p$\\
                $R_b = f(\mathcal{P}_p)$
            }
        }
        \uElse{
          $\mathcal{P}_p = \mathcal{P}_p  + [a]$ 
        }
        Sample a mini-batch of transition tuples from $\mathcal{M}$ \\
        Update $\mathcal{Q}$ with gradient descent. 
    }
    \uIf{$e \mod{K} = 0$}{
        Construct a path $\mathcal{P}$ greedily based on $\mathcal{Q}$\\
        Update $\mathcal{P}_b$  with $\mathcal{P}$ if $\mathcal{P}$ has a larger reward
    }
}             
       
return $\mathcal{P}_b$
\end{algorithm}
The learning and searching procedure is outlined in Algorithm~\ref{alg:ls}. More details in terms of deep Q-learning training techniques can be found in~\cite{mnih2013playing}.

\section{Performance Evaluation}
\label{sect:pe}
In this section, the performance of the proposed Q-learning approach for IPP is evaluated. In particular, we compare with a naive exploration approach in terms of learning efficiency and also compare the performance with other IPP algorithms. Finally, we show that the knowledge of the Q-network is transferable when the constraints change, especially in cases when the changes are moderate.

\subsection{Graph Setting and RL Implementation}
\begin{figure}[!t]
\centering
\includegraphics[width=0.4\textwidth]{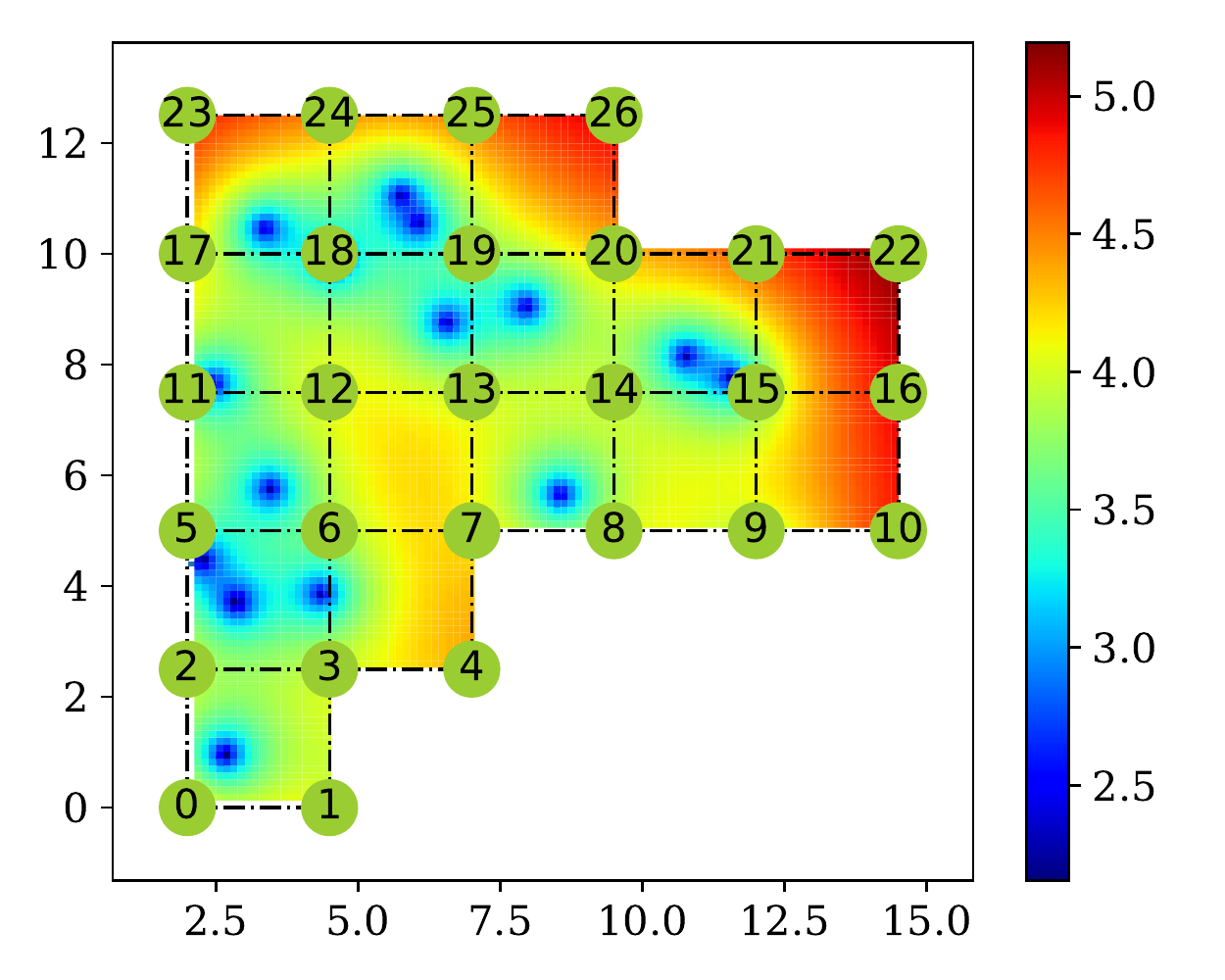}
\caption{Graph generated from Area One. The size of the whole area is approximately 12m * 13m. The X and Y axes show the dimensions in meters, and the color represents the uncertainty (entropy) of the predicted signals by fitting a GP with the pilot data.  The grid graph has 26 vertices.}
\label{fig:area1}
\end{figure}

\begin{figure}[!t]
\centering
\includegraphics[width=0.5\textwidth]{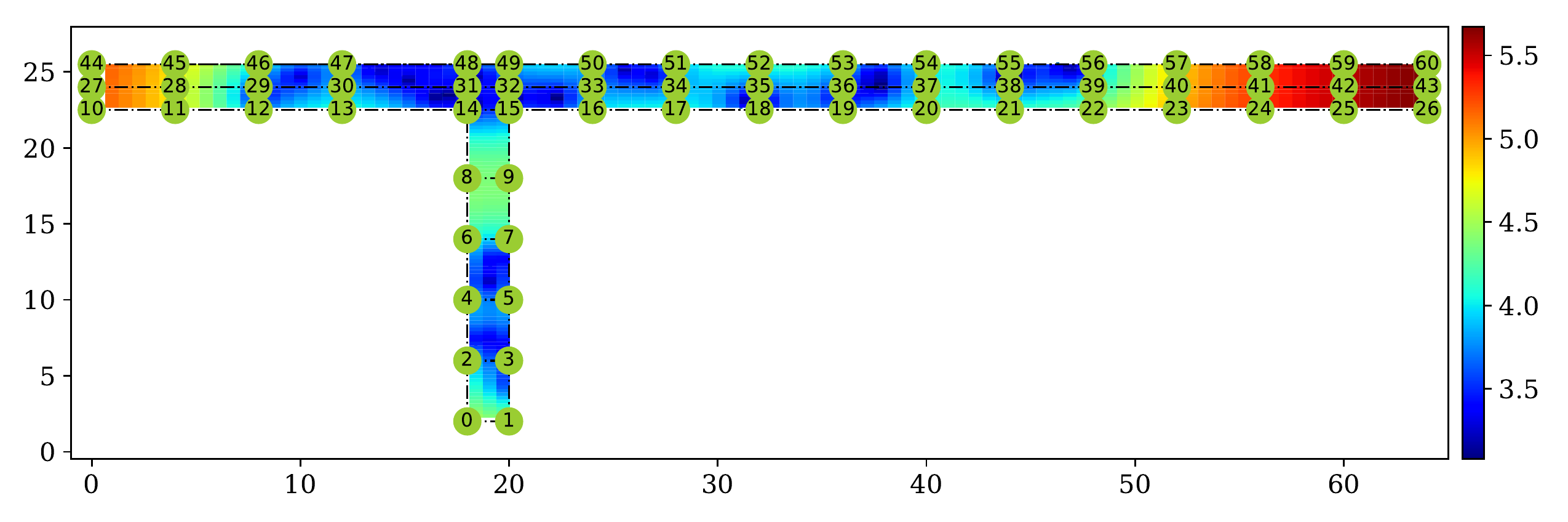}
\caption{Graph generated from Area Two. This area is a ``T" shape corridor, with 25m in height and 64m in length. The graph has 61 vertices as shown by the green circles.}
\label{fig:area2}
\end{figure}

In experiments, we consider WiFi signal strengths as the environmental data, which have been extensively used for fingerprint-based indoor localization~\cite{wu2012will,yang2012locating,li2017turf}. Two real-world indoor areas are selected and discretized into grid graphs. The first area is an open area and the second area is a corridor. A small amount of pilot WiFi signals are collected to estimate the hyperparamers of the underlying GP for each area. The two areas are illustrated in Fig.\ref{fig:area1} and Fig.\ref{fig:area2}, respectively.


A simulator representing the interactions between the agent and the graph is implemented with Python.  The APIs are similar to those in the OpenAI Gym~\cite{brockman2016openai}, which is a reinforcement learning platform.  For IPP, the main logic of the simulator is the state transition and reward mechanism as outlined in Algorithm~\ref{alg:strm}. 

The  RNN for Q-function is implemented in PyTorch, where each RNN cell is a LSTM unit. We adopt a double Q-learning~\cite{van2016deep} structure with prioritized experience replay~\cite{schaul2015prioritized}.

\subsection{Comparison with Naive Exploration}
\begin{figure}[!t]
	\centering
	\subfloat[][Naive Exploration]{\includegraphics[width=0.245\textwidth]{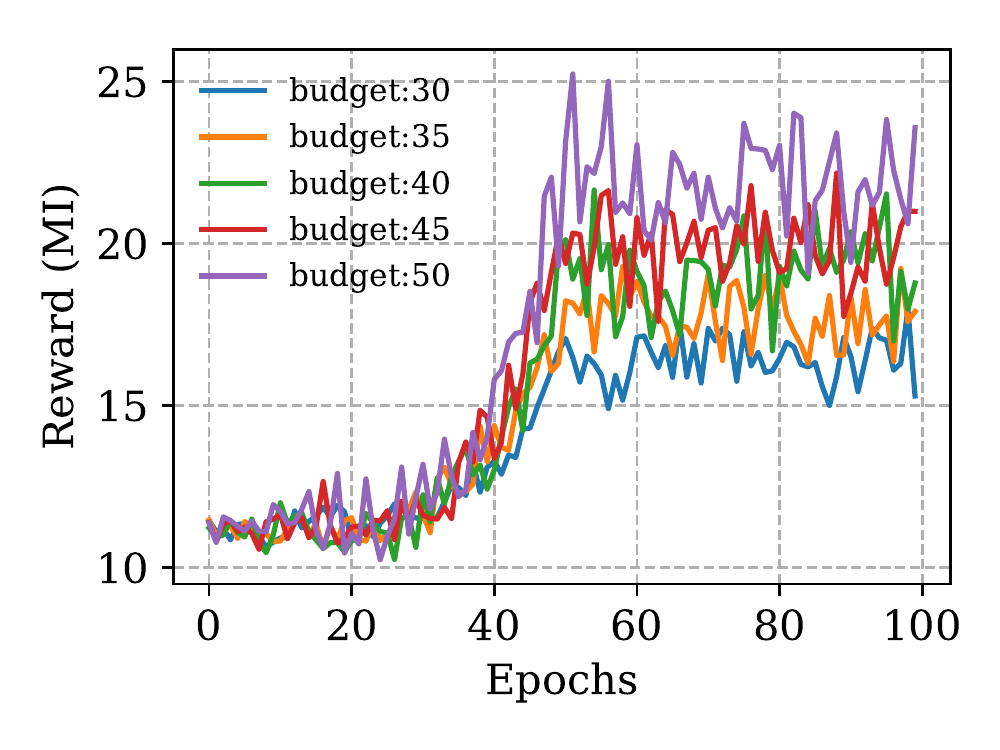} \label{fig:ns1}}%
	\subfloat[][Constrained Exploration]{\includegraphics[width=0.245\textwidth]{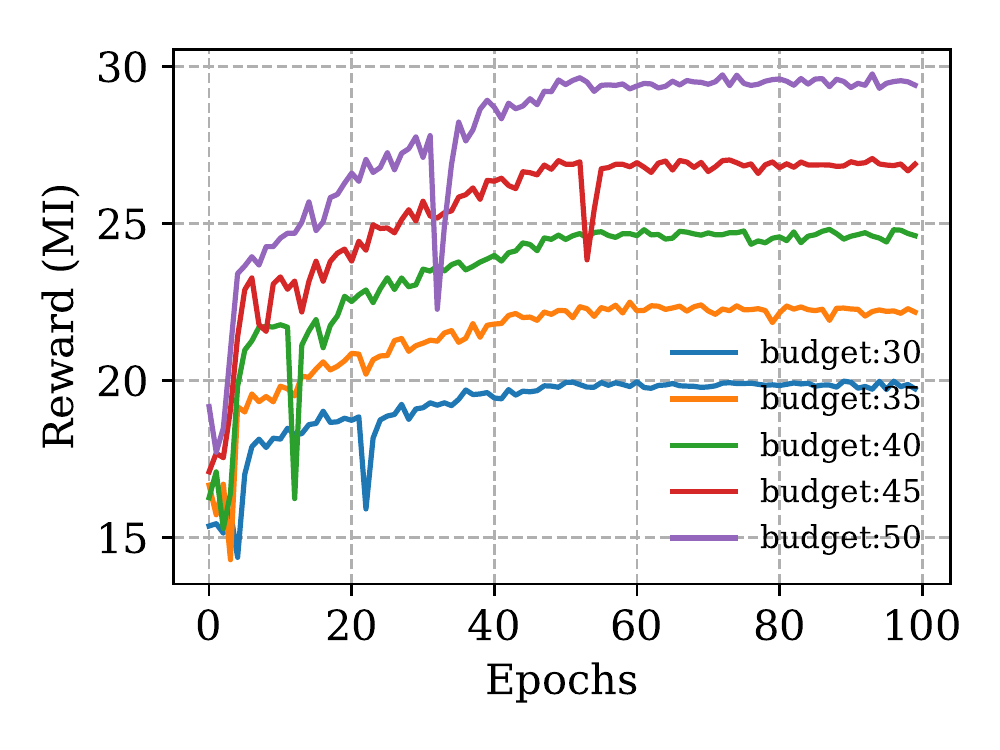} \label{fig:cs1}}%
	\caption{Average reward per episode with Q-learning in the graph from Area One. The start and terminal vertices are set to 0, so the path forms a tour. Experiments are run for different budgets (maximum distance) with $\epsilon$-greedy policy with $\epsilon=0.9$ initially and decay to $\epsilon=0.1$ at the 50th epoch. Each epoch means learning for 50 episodes, and the Y axis shows the average reward.  \protect\subref{fig:ns1} shows the naive exploration approach and \protect\subref{fig:cs1} shows the constrained exploration with shortest path.}
	\label{fig:area1epsreward}
\end{figure}

\begin{figure}[!t]
	\centering
	\subfloat[][Naive Exploration]{\includegraphics[width=0.245\textwidth]{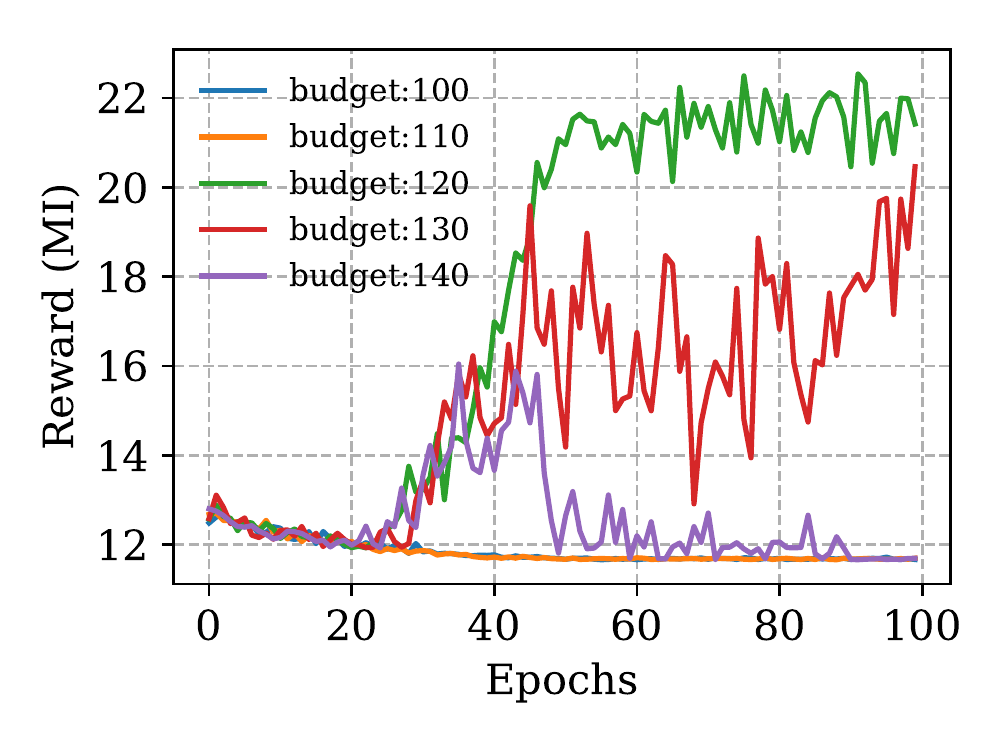} \label{fig:ns2}}%
	\subfloat[][Constrained Exploration]{\includegraphics[width=0.245\textwidth]{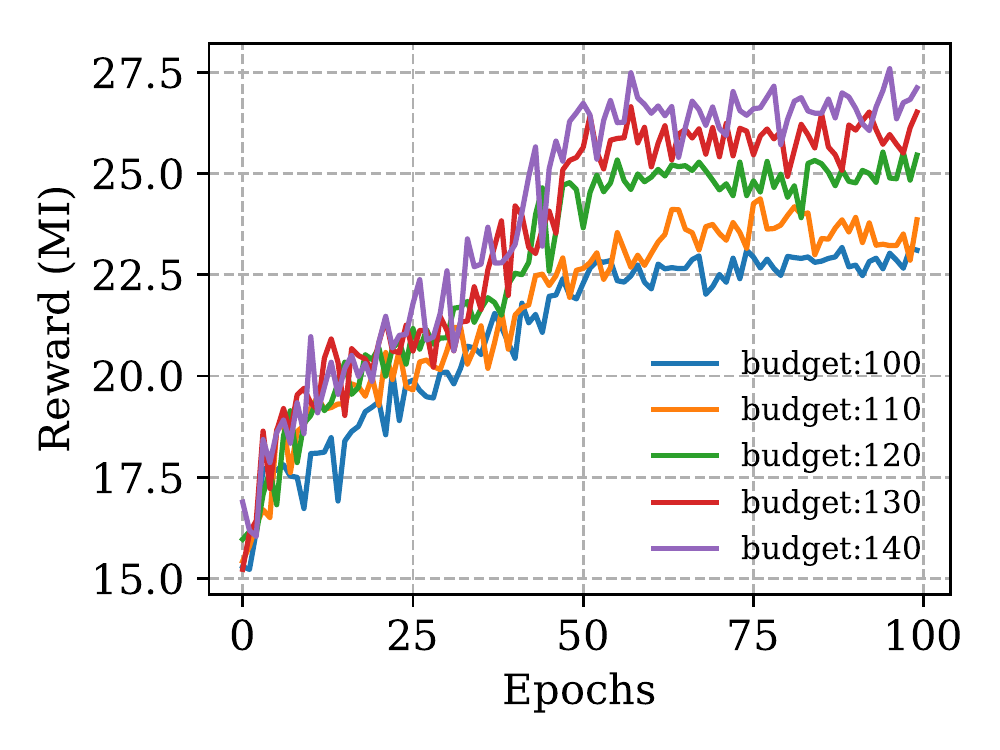} \label{fig:cs2}}%
	\caption{Average reward per episode with Q-learning in the graph from Area Two. The parameter settings are similar with Fig.\ref{fig:area1epsreward}.}
	\label{fig:area2epsreward}
\end{figure}

We first compare the performance with a naive exploration approach, which simply extends the partial path through neighborhood vertices until the budget is used up, and all the other settings are kept the same with the constrained exploration and exploitation strategy.  Fig.~\ref{fig:area1epsreward} and Fig.~\ref{fig:area2epsreward} show the average episode reward with the learning process in Area One and Two, respectively. Similar to~\cite{mnih2013playing}, each epoch is defined as 50 episodes of learning, and 100 epochs are run for each configuration. It can be seen clearly that the constrained exploration and exploitation strategy achieves higher reward (MI) and higher efficiency. During the initial episodes of the naive exploration, the rewards are low (penalized by $-1.0 \times R_c$ if fails to reach $v_t$, but not 0 because MI from the pilot data are considered) since most generated paths are invalid, i.e, not terminate at $v_t$. Furthermore, the difference is more significant in Area Two since the graph size is larger than Area One. In a larger graph, the probability that blind searches can construct a valid path is smaller. As can be seen from Fig.\ref{fig:area2epsreward}, for some budget setting (e.g., 100, 110, 140) the naive approach failed to improve in terms of average reward. In comparison, the constrained exploration and exploitation strategy shows a promising result, and the average reward improves gradually until convergence under different budget settings.

\subsection{Comparison with Other IPP algorithms}
The ultimate goal of IPP is to plan a path that can reduce prediction uncertainty with GP regression using the collected data. Unlike existing heuristics or evolution based approaches, the Q-learning solution learns from trial paths and improves gradually. For comparison, we have also implemented the following algorithms:

\paragraph*{Brute Force Tree Search} 
The brute force tree search tries to enumerate all the paths from $v_s$ to $v_t$ and record the path with the highest reward. A stack is utilized to store the partial paths and branches are searched similar to the depth-first-search traverse. Here $v_s$ can be seen as the root of the search tree. A search branch is terminated whether $v_t$ is encountered or budget is exhausted.

\paragraph*{Recursive Greedy Algorithm} 
The Recursive Greedy (RG) algorithm is adapted from~\cite{chekuri2005recursive}. Originally, RG is designed for the orienteering problem. For IPP, the reward function is adapted to consider samples along edges.

\paragraph*{Greedy Algorithm}
The greedy algorithm is implemented following~\cite{wei2018informative}. Vertices are selected greedily based on the marginal reward-cost ratio, and a Stainer TSP solver is implemented based on~\cite{applegate2006concorde} to generate paths since the graph is not complete.

\paragraph*{Genetic Algorithm}
Genetic Algorithm is implemented according to~\cite{wei2018informative}. Each valid path represents a chromosome, and a set of individuals (paths) are initialized. For each generation, a crossover and a mutation process are implemented. After a number of generations, the path with the maximum MI is considered as the final solution.

\begin{figure}[!t]
	\centering
	\subfloat[][Tour]{\includegraphics[width=0.245\textwidth]{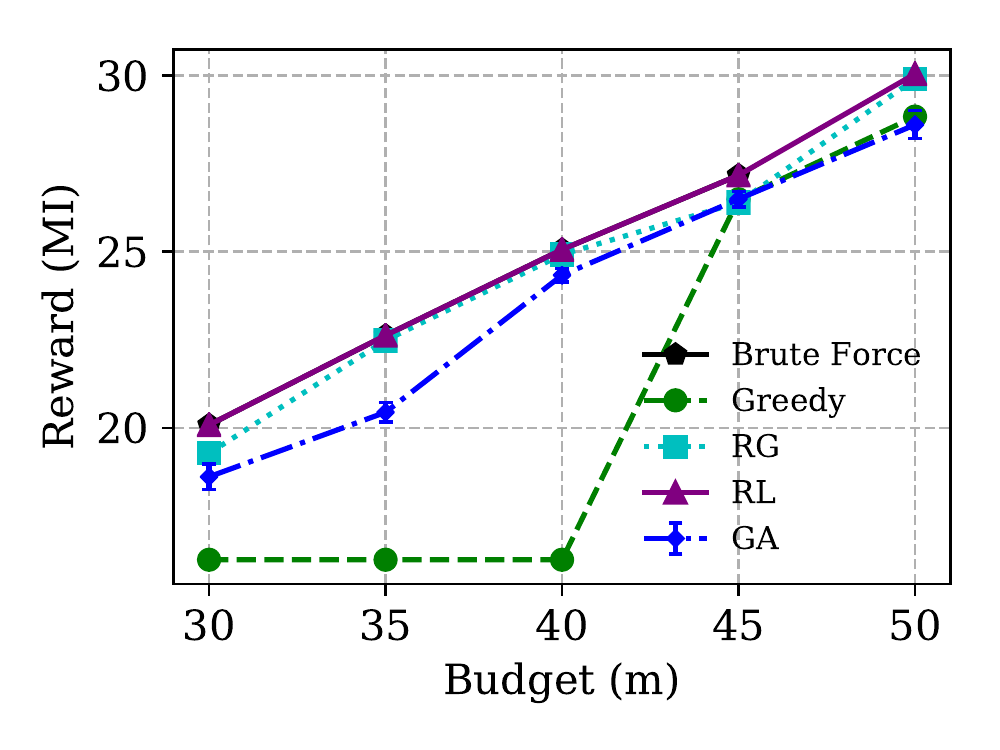} \label{fig:tu1}}%
	\subfloat[][Non-tour]{\includegraphics[width=0.245\textwidth]{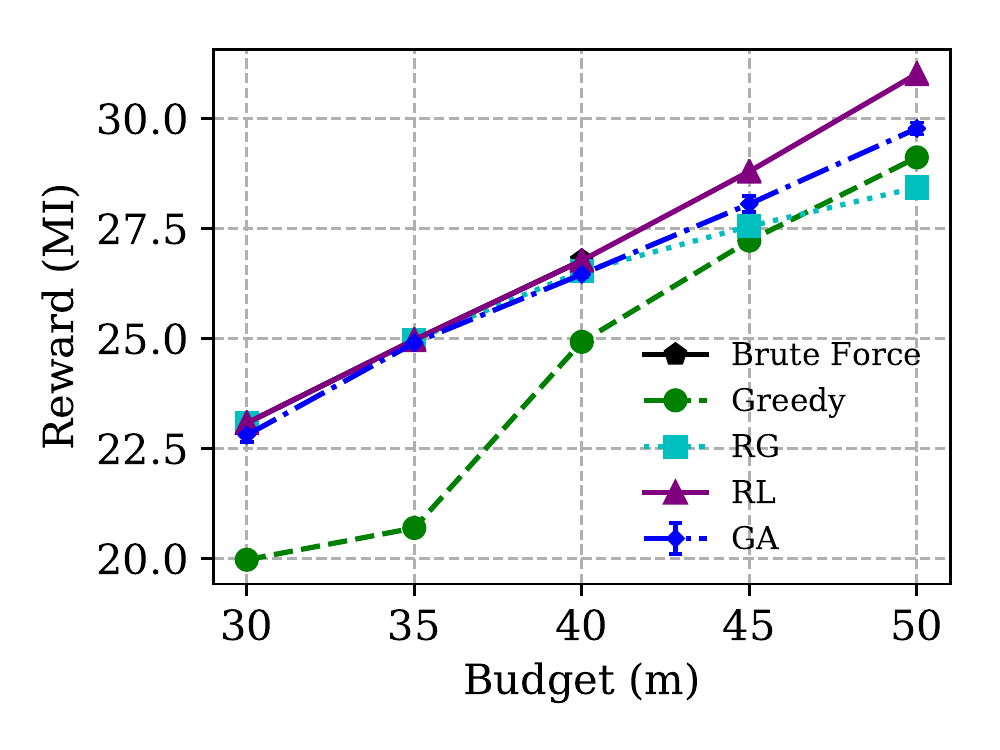} \label{fig:ntu1}}%
	\caption{Best path MI comparison with different algorithms for Area One. The start vertex $v_s$ is set to 0. For the non-tour case, the terminal vertex $v_t$ is set to 26. For RL, 5000 episodes are iterated, and for GA, the population size is set to 100 and 50 generations are iterated. The brute force approach is successful only when the budget are 30,35 and 40 given 72 hours of run time, please note that in the figure it is overlapped with RL.}
	\label{fig:area1utility}
\end{figure}

\begin{figure}[!t]
	\centering
	\subfloat[][Tour]{\includegraphics[width=0.245\textwidth]{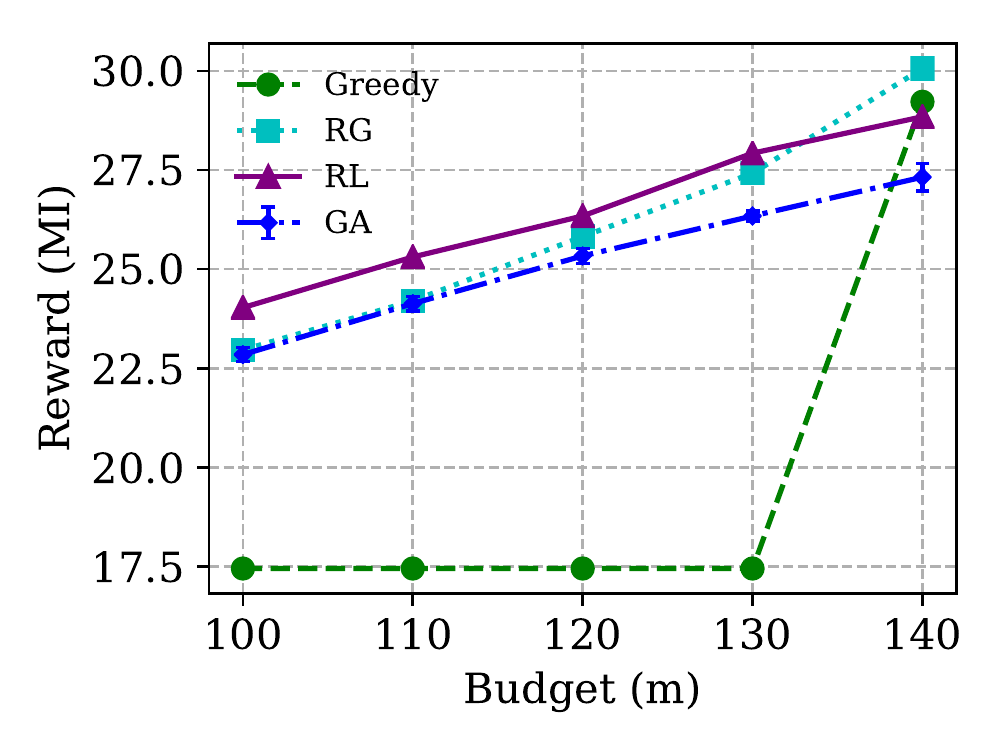} \label{fig:tu2}}%
	\subfloat[][Non-tour]{\includegraphics[width=0.245\textwidth]{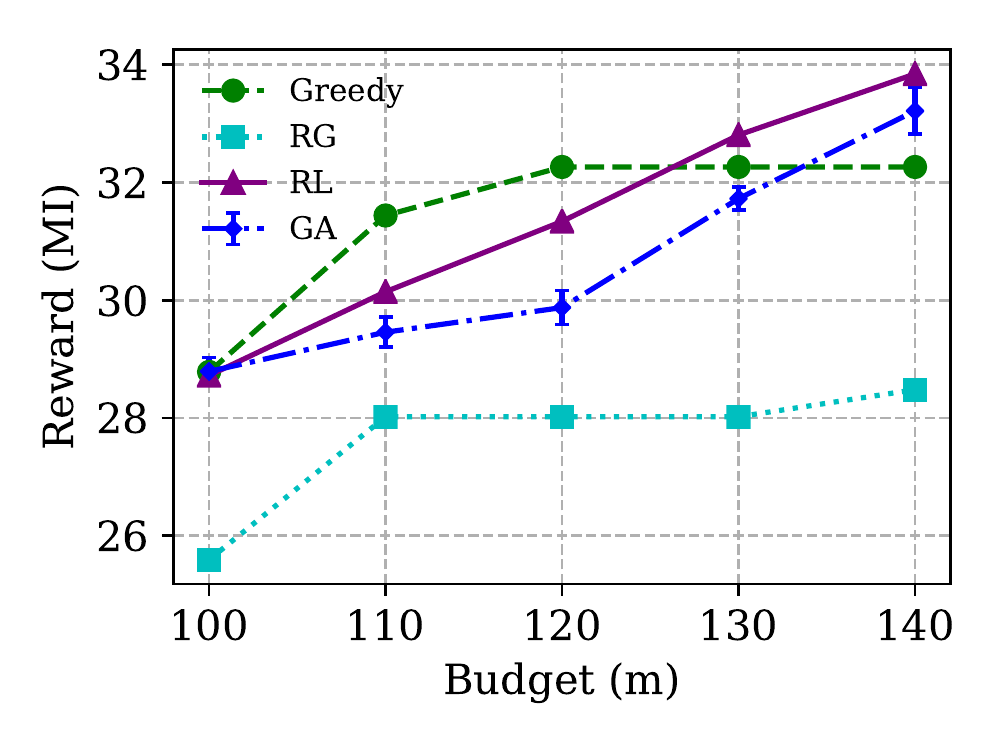} \label{fig:ntu2}}%
	\caption{Best path MI comparison with different algorithms for Area Two. The start vertex $v_s$ is set to 0. For the non-tour case, the terminal vertex $v_t$ is set to 60. For RL, 10000 episodes are iterated, and for GA, the population size is set to 100 and 100 generations are iterated.}
	\label{fig:area2utility}
\end{figure}

Both the brute force and RG approaches suffer from high computation complexity. The brute force approach is only applied to the graph of Area One, and it manages to return the result in 72 hours only when the budget is below 40 meters.

Both RL and GA are able to improve from trial paths, but accomplish so differently. RL is a learning based algorithm, while GA is an evolutionary algorithm. In RL, each trial path is an episode, the agent learns to make decisions for path construction. In contrast, in GA, the information is inherited through genetic operators such as cross-over and mutation, and each individual represents a trial path. For a fair comparison, we learn for 5000 episodes with RL in Area One. With GA, the population size is set to 100, and we run 50 generations. Thus, the total number of individuals (paths) involved are $100*50=5000$. Due to  randomness, we run five rounds of experiments independently and take the average for each budget setting. Similarly, for Area Two, the number of episodes for RL is set to 10000, and 100 generations are run for GA accordingly.

Meanwhile, for each Area, we consider both the tour case and a non-tour case. The tour case means the agent must return to the start vertex, i.e., $v_t = v_s$. While for the non-tour case, the terminal vertex is selected to be different from the start vertex.

It can be seen from Fig.~\ref{fig:area1utility} that RL achieves the best performance compared with all the other algorithms. When the budget is under 40 meters, the optimal solution can be found by RL, since they coincide with those from the brute force search. The rewards obtained by GA and RG increase monotonically with budgets, while the rewards from the greedy algorithm sometimes remain unchanged even if the budgets increase.

The graph from Area Two contains 61 vertices, with budgets larger than Area One.  Fig.~\ref{fig:area2utility} shows the results from RL, RG, GA and the greedy approach. RL outperforms the other algorithms for most of the budget settings. However, on the non-tour case in Fig.~\ref{fig:ntu2}, for two budget settings (110, 120), the greedy approach achieves the best results. 

\subsection{Transfer Learning}
In practice, the budget constraint $B$ depends on battery capacity.  Meanwhile, the start and terminal vertices could change if the locations of the charging stations change. One natural question is whether it is possible to adapt the trained models when these constraints change. Specifically, the parameters of the Q-network can be initialized randomly or  initialized from pre-trained models, this is known as transfer learning. In this section, experiments are carried out to demonstrate that the learned models are transferable when one of the constraints changes. 

\begin{figure}[!t]
	\centering
	\subfloat[][Area One]{\includegraphics[width=0.245\textwidth]{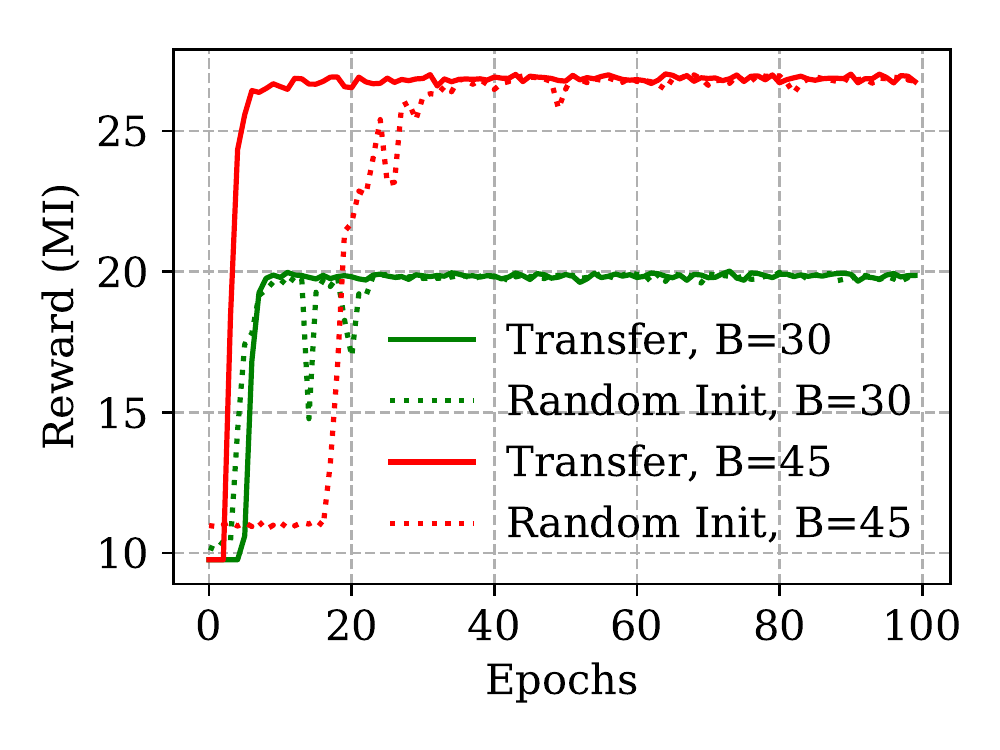} \label{fig:area1diffb}}%
	\subfloat[][Area Two]{\includegraphics[width=0.245\textwidth]{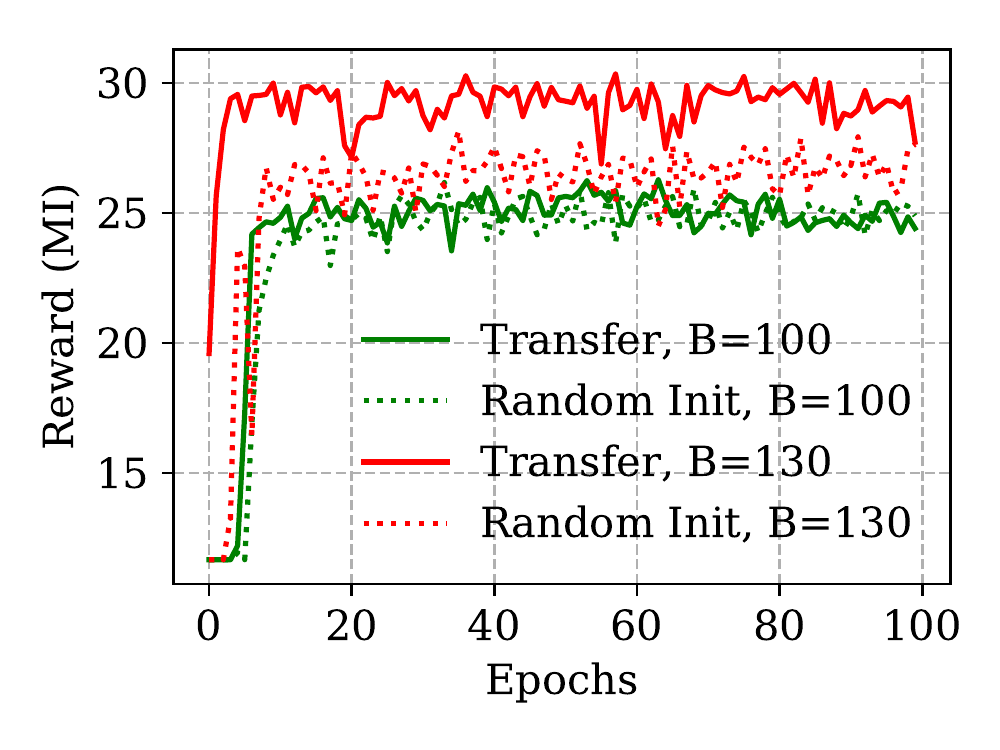} \label{fig:area2diffb}}%
	\caption{Transfer learning with different budgets. For Area One, $v_s=0, v_t=0$, the base model is trained with $v_s = 0,v_t=0, B = 50$. For Area Two, $v_s=0, v_t=0$, the base model is trained with $v_s=0,v_t=0, B=140$.}
	\label{fig:transdiffb}
\end{figure}

\begin{figure}[!t]
	\centering
	\subfloat[][Area One]{\includegraphics[width=0.245\textwidth]{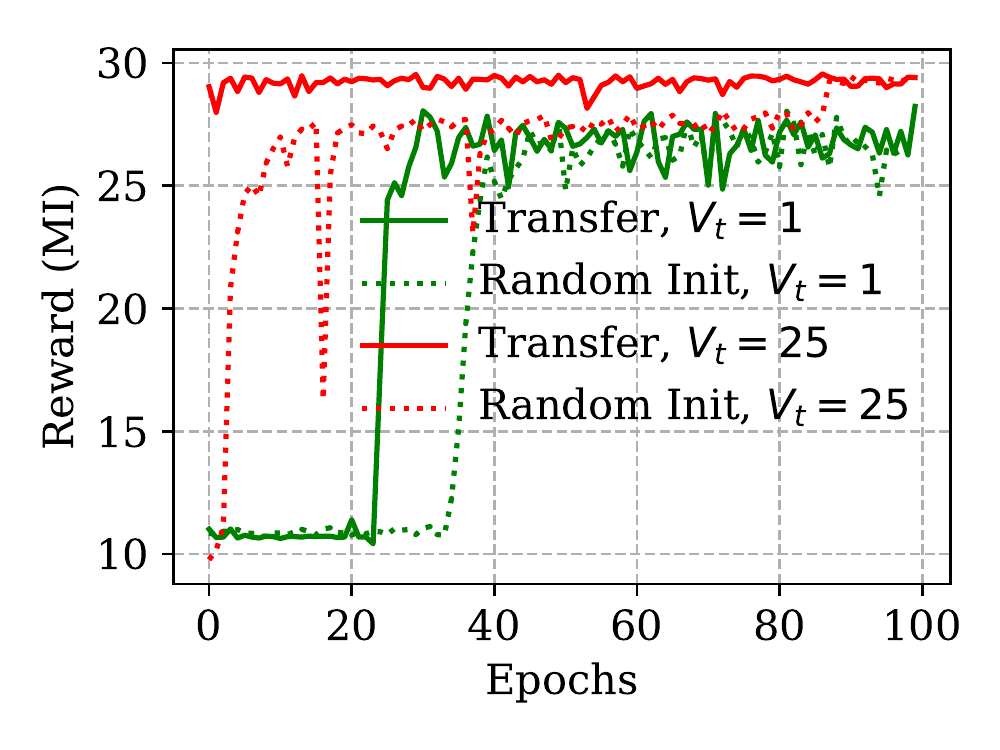} \label{fig:area1diffvt}}%
	\subfloat[][Area Two]{\includegraphics[width=0.245\textwidth]{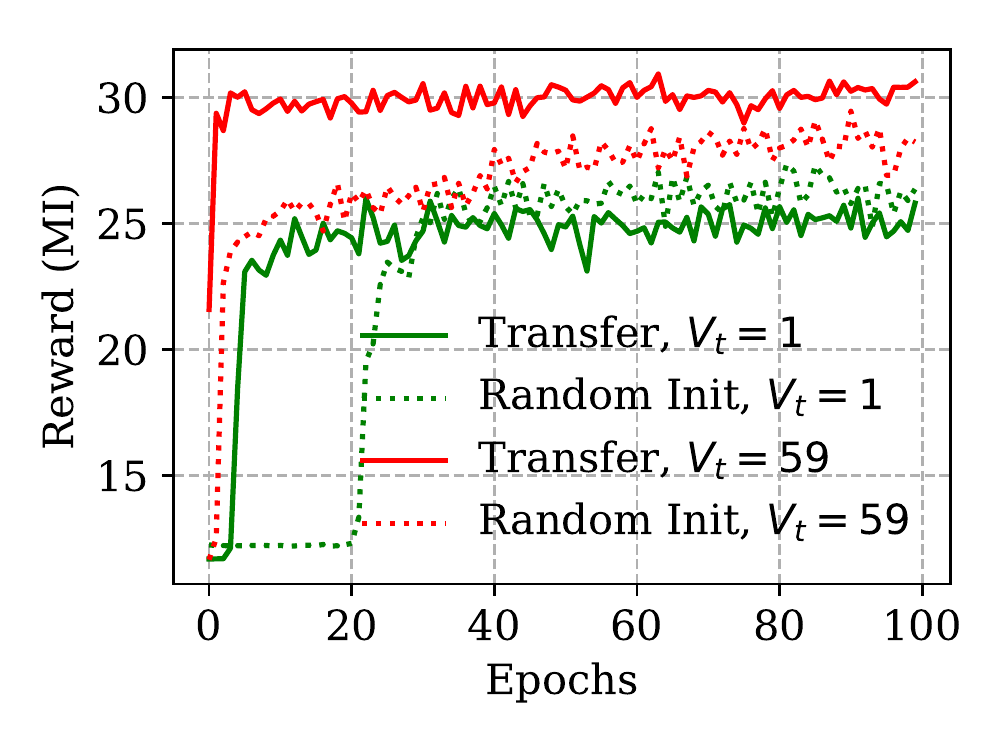} \label{fig:area2diffvt}}%
	\caption{Transfer learning with different terminal vertices.  For Area One, $v_s=0, B= 50$, the base model is trained with $v_s = 0,v_t=26, B = 50$. For Area Two, $v_s=0, B = 140$, the base model is trained with $v_s=0,v_t=60, B=140$.}
	\label{fig:transdiffvt}
\end{figure}

\begin{figure}[!t]
	\centering
	\subfloat[][Area One]{\includegraphics[width=0.245\textwidth]{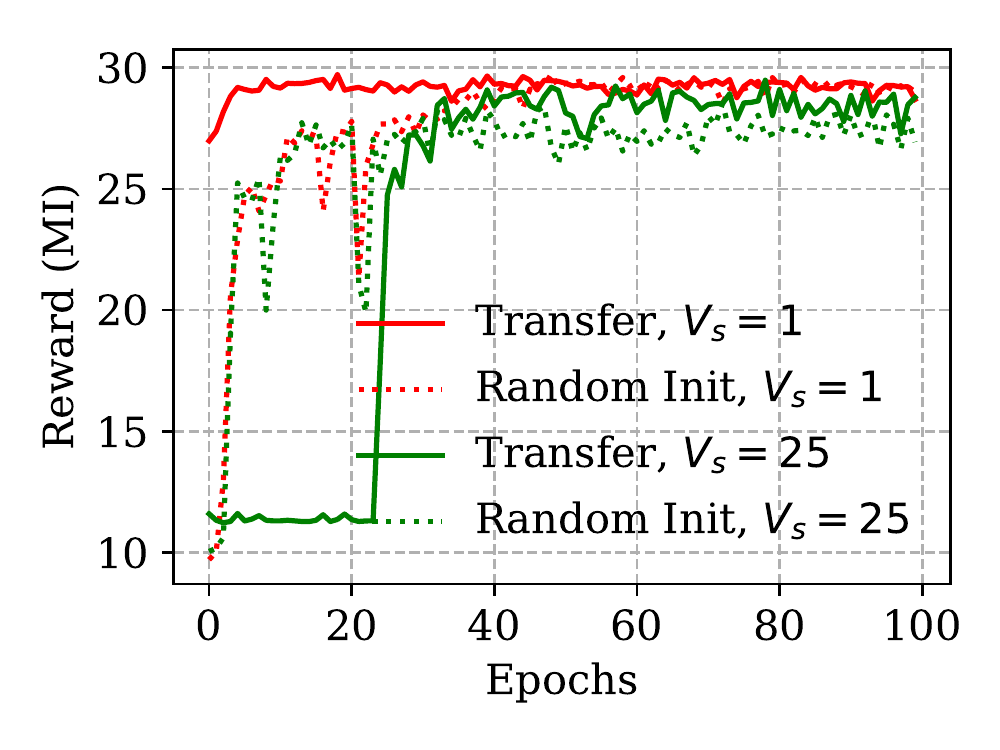} \label{fig:area1diffvs}}%
	\subfloat[][Area Two]{\includegraphics[width=0.245\textwidth]{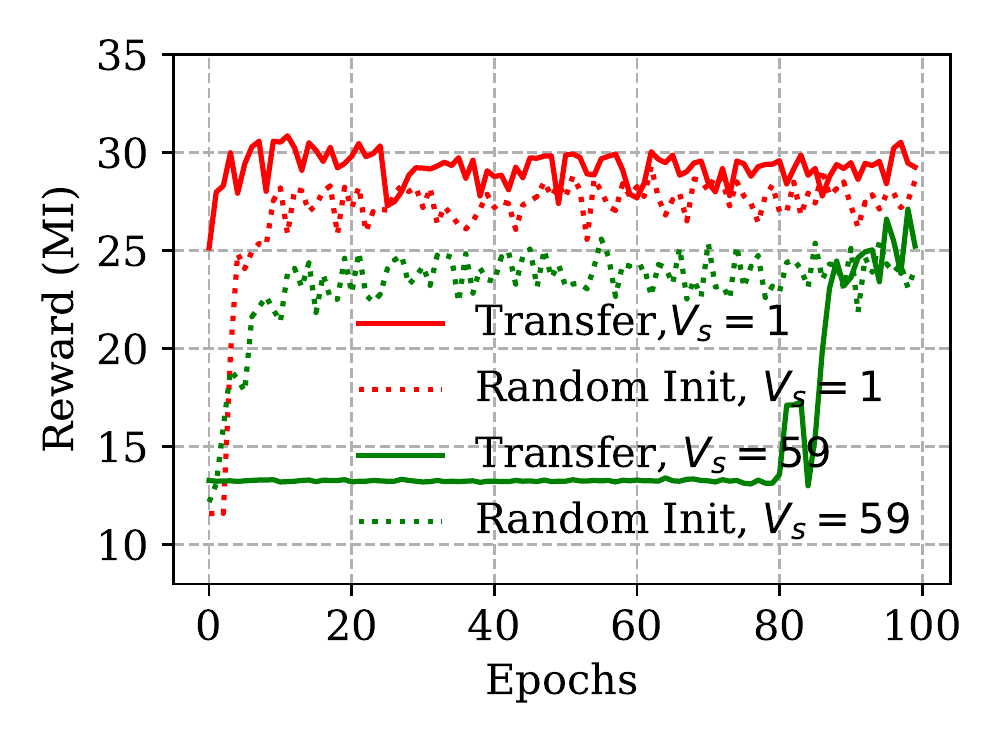} \label{fig:area2diffvs}}%
	\caption{Transfer learning with different start vertices.  For Area One, $v_t=26, B= 50$, the base model is trained with $v_s = 0,v_t=26, B = 50$. For Area Two, $v_t=60, B = 140$, the base model is trained with $v_s=0,v_t=60, B=140$.}
	\label{fig:transdiffvs}
\end{figure}

\paragraph*{Different Budget} Fig.~\ref{fig:transdiffb} shows the effect of transfer learning when the budget changes. For each area, a base model is learned first. Then we change the budget, the learning curves of random initialization and fine tune from the base model are compared. It can be seen that in both areas when the budget is close to the base model, the effect of transfer learning is clear since the model converges faster. When the budget is far from the base model, the advantage is less significant.
\paragraph*{Varying Terminal Vertices} Fig.~\ref{fig:transdiffvt} shows the transfer learning effect when the terminal vertex changes, and the start and budget keep the same. In both areas, transfer learning shows a earlier convergence time compared with random initialization.
\paragraph*{Varying Start Vertices}Fig.~\ref{fig:transdiffvs} shows the result when the start vertex is changed. In both areas, we observed that when the new start vertex is close to the start vertex from the base model, transfer learning is advantageous. However, when the start vertex is far apart from that in the pre-trained model, random initialization performs better. 

From the above comparison we can conclude that the learned models are transferable, particularly when the changes ($B,v_s,v_t$) are moderate. This can be attributed to the fact that the Q-network is learned from the transition tuples stored in the experience buffer. When the constraints are similar or close, the experience buffer tends to have identical transition tuples. Thus, model parameters are expected to be adapted using less transition tuples.

\subsection{Computation Complexity}

The RG suffers from a high computation complexity with $O((2nB)^I \cdot T_f)$~\cite{chekuri2005recursive}, where $n$ is the number of vertices and $T_f$ is the maximum time to evaluate the reward function on a given set of vertices, and $I$ is the recursion depth. The Greedy algorithm relies on the TSP solver to generate paths, and the complexity can be expressed as $O(Bn * t(n))$, where $t(n)$ is the complexity of the adopted TSP solver.

GA is an evolutionary algorithm, and the complexity is dominated by the defined number of generations and population size. Similarly, RL is a learning based algorithm and its complexity depends on the number of episode iterated and the budget size, since more budget means for within each episode there are more steps.

\begin{figure}[!t]
	\centering
	\subfloat[][Area One]{\includegraphics[width=0.245\textwidth]{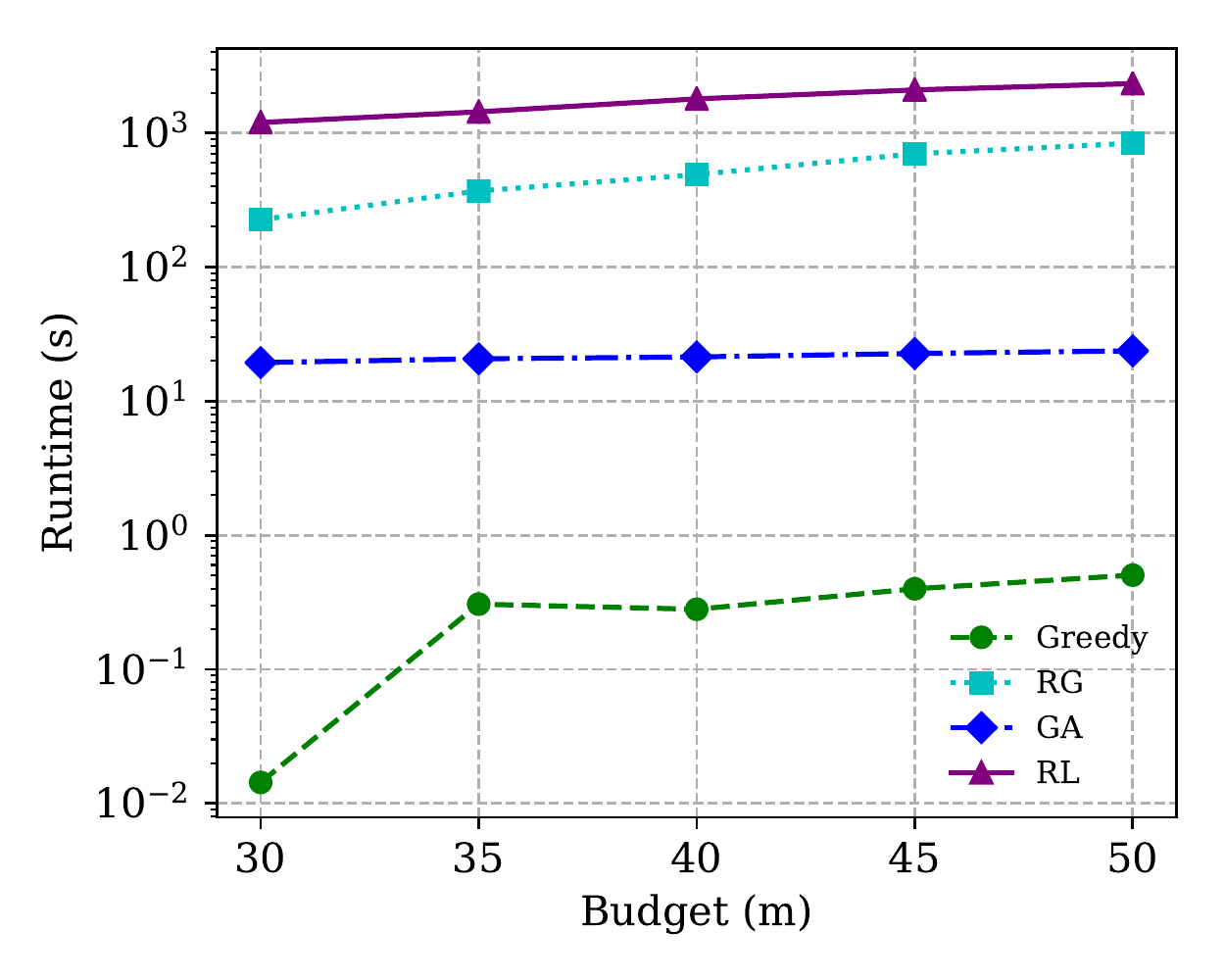} \label{fig:area1rt}}%
	\subfloat[][Area Two]{\includegraphics[width=0.245\textwidth]{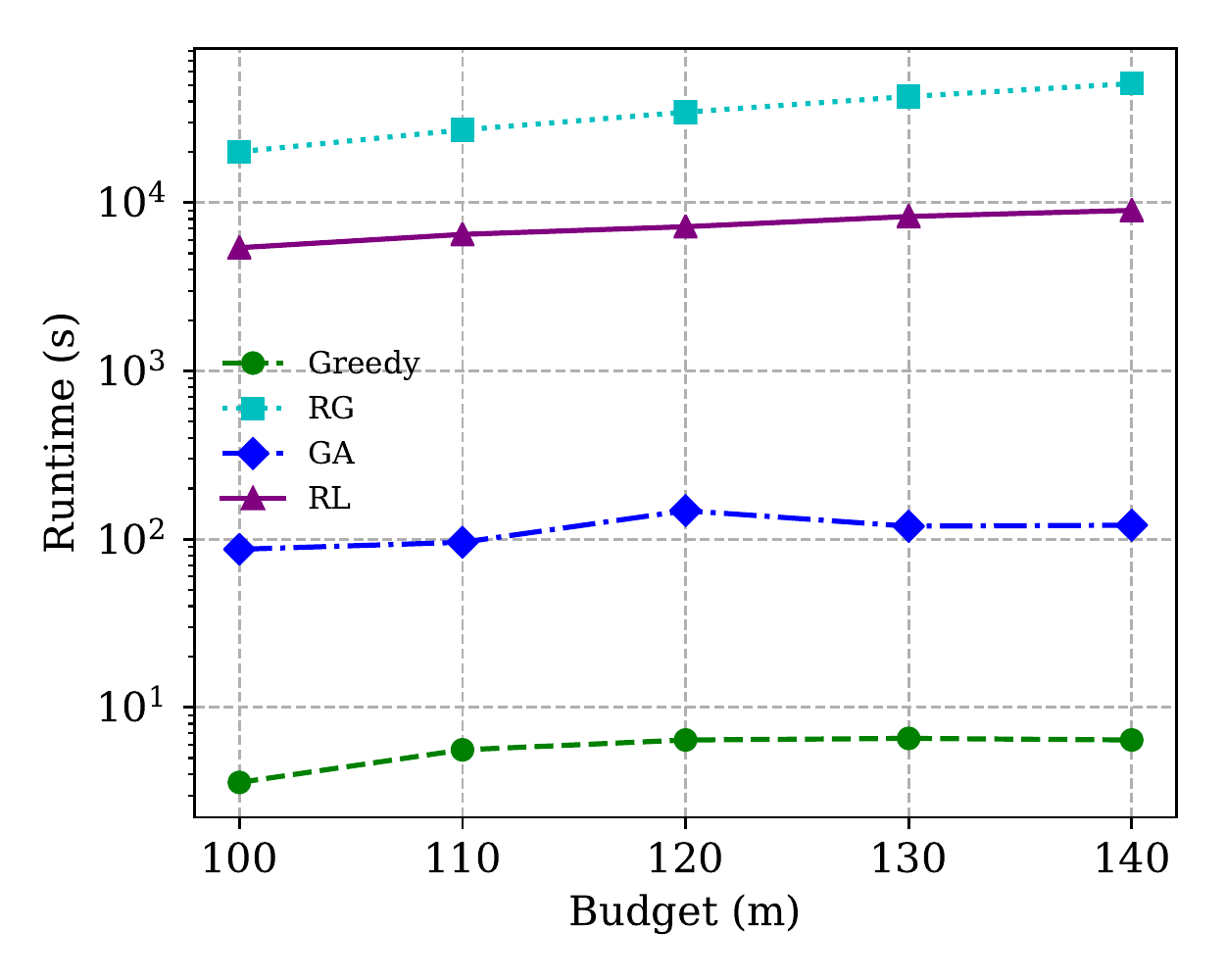} \label{fig:area2rt}}%
	\caption{Approximate execution time of different algorithms for the graph of Area One and Two on iMac (4GHz, Intel Core i7). For Area One, RL is run for 5000 episodes, and GA is iterated for 50 generations with a population size of 100.  For Area Two, RL is run for 10000 episodes, and GA is iterated for 100 generations with a population size of 100. The recursion depth $I$ of RG is set to two in both cases.}
	\label{fig:runtime}
\end{figure}

The execution time on an iMac desktop computer (4GHz Intel Core i7, 16 GB RAM, without GPU) is shown in Fig.~\ref{fig:runtime}. In general, GA and the Greedy algorithm is fast and can finish execution within a few minutes. Due to the training and optimization of neural network, RL takes a longer time than RG on the small graph (Area One). However, the execution time of RG increases exponentially when the number of nodes $n$ and budgets $B$ increase. In contrast, the execution time of RL grows linearly with budget and the number of iterations, and thus in Area Two RL takes less time than RG.

\section{Conclusion}
\label{sect:conclustion}
In this paper, a Q-learning based solution to IPP was presented. We proposed a novel exploration and exploitation strategy with the assistance of the shortest path. Compared with the naive exploration strategy, it has a better efficiency and optimality. Furthermore, the result is promising compared with state-of-the-art algorithms. We also demonstrated that the Q-network is transferable in presence of moderate changes in the input parameters. Our future research direction is to investigate the IPP problem for multiple robots.

\bibliographystyle{IEEEtran}
\bibliography{ref}

\begin{thebibliography}{10}
\providecommand{\url}[1]{#1}
\csname url@samestyle\endcsname
\providecommand{\newblock}{\relax}
\providecommand{\bibinfo}[2]{#2}
\providecommand{\BIBentrySTDinterwordspacing}{\spaceskip=0pt\relax}
\providecommand{\BIBentryALTinterwordstretchfactor}{4}
\providecommand{\BIBentryALTinterwordspacing}{\spaceskip=\fontdimen2\font plus
\BIBentryALTinterwordstretchfactor\fontdimen3\font minus
  \fontdimen4\font\relax}
\providecommand{\BIBforeignlanguage}[2]{{%
\expandafter\ifx\csname l@#1\endcsname\relax
\typeout{** WARNING: IEEEtran.bst: No hyphenation pattern has been}%
\typeout{** loaded for the language `#1'. Using the pattern for}%
\typeout{** the default language instead.}%
\else
\language=\csname l@#1\endcsname
\fi
#2}}
\providecommand{\BIBdecl}{\relax}
\BIBdecl

\bibitem{akyildiz2002wireless}
I.~F. Akyildiz, W.~Su, Y.~Sankarasubramaniam, and E.~Cayirci, ``Wireless sensor
  networks: a survey,'' \emph{Computer networks}, vol.~38, no.~4, pp. 393--422,
  2002.

\bibitem{di2011data}
M.~Di~Francesco, S.~K. Das, and G.~Anastasi, ``Data collection in wireless
  sensor networks with mobile elements: A survey,'' \emph{ACM Transactions on
  Sensor Networks (TOSN)}, vol.~8, no.~1, p.~7, 2011.

\bibitem{wang2007robot}
Y.~Wang and C.-H. Wu, ``Robot-assisted sensor network deployment and data
  collection,'' in \emph{2007 International Symposium on Computational
  Intelligence in Robotics and Automation}.\hskip 1em plus 0.5em minus
  0.4em\relax IEEE, 2007, pp. 467--472.

\bibitem{williams2006gaussian}
C.~K. Williams and C.~E. Rasmussen, \emph{Gaussian processes for machine
  learning}.\hskip 1em plus 0.5em minus 0.4em\relax MIT Press Cambridge, MA,
  2006, vol.~2, no.~3.

\bibitem{guestrin2005near}
C.~Guestrin, A.~Krause, and A.~P. Singh, ``Near-optimal sensor placements in
  gaussian processes,'' in \emph{Proceedings of the 22nd international
  conference on Machine learning}.\hskip 1em plus 0.5em minus 0.4em\relax ACM,
  2005, pp. 265--272.

\bibitem{singh2007efficient}
A.~Singh, A.~Krause, C.~Guestrin, W.~J. Kaiser, and M.~A. Batalin, ``Efficient
  planning of informative paths for multiple robots.'' in \emph{IJCAI}, vol.~7,
  2007, pp. 2204--2211.

\bibitem{binney2010informative}
J.~Binney, A.~Krause, and G.~S. Sukhatme, ``Informative path planning for an
  autonomous underwater vehicle,'' in \emph{2010 IEEE International Conference
  on Robotics and Automation}.\hskip 1em plus 0.5em minus 0.4em\relax IEEE,
  2010, pp. 4791--4796.

\bibitem{meliou2007nonmyopic}
A.~Meliou, A.~Krause, C.~Guestrin, and J.~M. Hellerstein, ``Nonmyopic
  informative path planning in spatio-temporal models,'' in \emph{AAAI},
  vol.~10, no.~4, 2007, pp. 16--7.

\bibitem{binney2012branch}
J.~Binney and G.~S. Sukhatme, ``Branch and bound for informative path
  planning,'' in \emph{2012 IEEE International Conference on Robotics and
  Automation}.\hskip 1em plus 0.5em minus 0.4em\relax IEEE, 2012, pp.
  2147--2154.

\bibitem{vansteenwegen2011orienteering}
P.~Vansteenwegen, W.~Souffriau, and D.~Van~Oudheusden, ``The orienteering
  problem: A survey,'' \emph{European Journal of Operational Research}, vol.
  209, no.~1, pp. 1--10, 2011.

\bibitem{yu2014informative}
S.~Yu, J.~Hao, B.~Zhang, and C.~Li, ``Informative mobility scheduling for
  mobile data collector in wireless sensor networks,'' in \emph{2014 IEEE
  Global Communications Conference}.\hskip 1em plus 0.5em minus 0.4em\relax
  IEEE, 2014, pp. 5002--5007.

\bibitem{wei2018informative}
Y.~Wei, C.~Frincu, and R.~Zheng, ``Informative path planning for location
  fingerprint collection,'' \emph{IEEE Transactions on Network Science and
  Engineering}, 2019.

\bibitem{viseras2016planning}
A.~Viseras, R.~O. Losada, and L.~Merino, ``Planning with ants: Efficient path
  planning with rapidly exploring random trees and ant colony optimization,''
  \emph{International Journal of Advanced Robotic Systems}, vol.~13, no.~5, p.
  1729881416664078, 2016.

\bibitem{wu2012will}
C.~Wu, Z.~Yang, Y.~Liu, and W.~Xi, ``Will: Wireless indoor localization without
  site survey,'' \emph{IEEE Transactions on Parallel and Distributed Systems},
  vol.~24, no.~4, pp. 839--848, 2012.

\bibitem{yang2012locating}
Z.~Yang, C.~Wu, and Y.~Liu, ``Locating in fingerprint space: wireless indoor
  localization with little human intervention,'' in \emph{Proceedings of the
  18th annual international conference on Mobile computing and
  networking}.\hskip 1em plus 0.5em minus 0.4em\relax ACM, 2012, pp. 269--280.

\bibitem{li2017turf}
C.~Li, Q.~Xu, Z.~Gong, and R.~Zheng, ``Turf: Fast data collection for
  fingerprint-based indoor localization,'' in \emph{2017 International
  Conference on Indoor Positioning and Indoor Navigation (IPIN)}.\hskip 1em
  plus 0.5em minus 0.4em\relax IEEE, 2017, pp. 1--8.

\bibitem{macdonald2019active}
R.~A. MacDonald and S.~L. Smith, ``Active sensing for motion planning in
  uncertain environments via mutual information policies,'' \emph{The
  International Journal of Robotics Research}, vol.~38, no. 2-3, pp. 146--161,
  2019.

\bibitem{arora2017randomized}
S.~Arora and S.~Scherer, ``Randomized algorithm for informative path planning
  with budget constraints,'' in \emph{Robotics and Automation (ICRA), 2017 IEEE
  International Conference on}.\hskip 1em plus 0.5em minus 0.4em\relax IEEE,
  2017, pp. 4997--5004.

\bibitem{ma2017informative}
K.-C. Ma, L.~Liu, and G.~S. Sukhatme, ``Informative planning and online
  learning with sparse gaussian processes,'' in \emph{2017 IEEE International
  Conference on Robotics and Automation (ICRA)}.\hskip 1em plus 0.5em minus
  0.4em\relax IEEE, 2017, pp. 4292--4298.

\bibitem{chekuri2005recursive}
C.~Chekuri and M.~Pal, ``A recursive greedy algorithm for walks in directed
  graphs,'' in \emph{46th Annual IEEE Symposium on Foundations of Computer
  Science (FOCS'05)}.\hskip 1em plus 0.5em minus 0.4em\relax IEEE, 2005, pp.
  245--253.

\bibitem{kaelbling1996reinforcement}
L.~P. Kaelbling, M.~L. Littman, and A.~W. Moore, ``Reinforcement learning: A
  survey,'' \emph{Journal of artificial intelligence research}, vol.~4, pp.
  237--285, 1996.

\bibitem{sutton2018reinforcement}
R.~S. Sutton and A.~G. Barto, \emph{Reinforcement learning: An
  introduction}.\hskip 1em plus 0.5em minus 0.4em\relax MIT press, 2018.

\bibitem{mnih2013playing}
V.~Mnih, K.~Kavukcuoglu, D.~Silver, A.~Graves, I.~Antonoglou, D.~Wierstra, and
  M.~Riedmiller, ``Playing atari with deep reinforcement learning,''
  \emph{arXiv preprint arXiv:1312.5602}, 2013.

\bibitem{van2016deep}
H.~Van~Hasselt, A.~Guez, and D.~Silver, ``Deep reinforcement learning with
  double q-learning,'' in \emph{Thirtieth AAAI Conference on Artificial
  Intelligence}, 2016.

\bibitem{wang2015dueling}
Z.~Wang, T.~Schaul, M.~Hessel, H.~Van~Hasselt, M.~Lanctot, and N.~De~Freitas,
  ``Dueling network architectures for deep reinforcement learning,''
  \emph{arXiv preprint arXiv:1511.06581}, 2015.

\bibitem{schaul2015prioritized}
T.~Schaul, J.~Quan, I.~Antonoglou, and D.~Silver, ``Prioritized experience
  replay,'' \emph{arXiv preprint arXiv:1511.05952}, 2015.

\bibitem{bello2016neural}
I.~Bello, H.~Pham, Q.~V. Le, M.~Norouzi, and S.~Bengio, ``Neural combinatorial
  optimization with reinforcement learning,'' \emph{arXiv preprint
  arXiv:1611.09940}, 2016.

\bibitem{khalil2017learning}
E.~Khalil, H.~Dai, Y.~Zhang, B.~Dilkina, and L.~Song, ``Learning combinatorial
  optimization algorithms over graphs,'' in \emph{Advances in Neural
  Information Processing Systems}, 2017, pp. 6348--6358.

\bibitem{golden1987orienteering}
B.~L. Golden, L.~Levy, and R.~Vohra, ``The orienteering problem,'' \emph{Naval
  Research Logistics (NRL)}, vol.~34, no.~3, pp. 307--318, 1987.

\bibitem{gunawan2016orienteering}
A.~Gunawan, H.~C. Lau, and P.~Vansteenwegen, ``Orienteering problem: A survey
  of recent variants, solution approaches and applications,'' \emph{European
  Journal of Operational Research}, vol. 255, no.~2, pp. 315--332, 2016.

\bibitem{yu2014correlated}
J.~Yu, M.~Schwager, and D.~Rus, ``Correlated orienteering problem and its
  application to informative path planning for persistent monitoring tasks,''
  in \emph{2014 IEEE/RSJ International Conference on Intelligent Robots and
  Systems}.\hskip 1em plus 0.5em minus 0.4em\relax IEEE, 2014, pp. 342--349.

\bibitem{cao2013multi}
N.~Cao, K.~H. Low, and J.~M. Dolan, ``Multi-robot informative path planning for
  active sensing of environmental phenomena: A tale of two algorithms,'' in
  \emph{Proceedings of the 2013 international conference on Autonomous agents
  and multi-agent systems}.\hskip 1em plus 0.5em minus 0.4em\relax
  International Foundation for Autonomous Agents and Multiagent Systems, 2013,
  pp. 7--14.

\bibitem{ahmed1989entropy}
N.~A. Ahmed and D.~Gokhale, ``Entropy expressions and their estimators for
  multivariate distributions,'' \emph{IEEE Transactions on Information Theory},
  vol.~35, no.~3, pp. 688--692, 1989.

\bibitem{brockman2016openai}
G.~Brockman, V.~Cheung, L.~Pettersson, J.~Schneider, J.~Schulman, J.~Tang, and
  W.~Zaremba, ``Openai gym,'' \emph{arXiv preprint arXiv:1606.01540}, 2016.

\bibitem{applegate2006concorde}
D.~Applegate, R.~Bixby, V.~Chvatal, and W.~Cook, ``Concorde tsp solver,'' 2006.

\end{thebibliography}
\end{document}